\documentclass{article}
\usepackage{amsmath,amssymb}
\usepackage{graphicx}
\usepackage{array}
\usepackage{wrapfig}
\usepackage[final]{corl_2026} 

\title{S2A2: Audio-Visual Imitation Learning for Manipulation Tasks Using Acoustic Spatial Information}

\author{
  Kaneyoshi Hiratsuka\\
  Kyoto University, RIKEN \\
  \AND
  Benjamin Yen \\
  Institute of Science Tokyo, RIKEN \\
  \And
  Ryosuke Kojima \\
  Kyoto University, RIKEN \\
}

\begin{document}
\maketitle
\begin{abstract}

Acoustic information provides rich cues about object location, material properties, and changes caused by contact or motion.
This paper introduces a new set of acoustic-aware manipulation tasks for imitation learning, in which robots must use auditory cues to determine manipulation targets.
These tasks require sound source localization and identification for active exploration in robotic manipulation.
Also, we propose a multimodal imitation learning framework, Spatial-Spectral Audio Action (S2A2), that integrates visual features with acoustic spatial and acoustic signal information for the acoustic-aware manipulation tasks.
We implemented S2A2 models that integrate policies such as ACT, Diffusion Policy, VQ-BeT, and $\pi_0$, into our framework. Simulation experiments showed that the proposed method is the most effective for tasks requiring both position and timbre.
Furthermore, real robot experiments confirm the applicability of the proposed tasks and framework to real-world manipulation.

\end{abstract}

\section{Introduction}
Imitation learning, which acquires manipulation policies from human demonstrations, has been widely studied as a framework for robots to acquire complex manipulation skills~\citep{pomerleau1988alvinn,argall2009survey,osa2018algorithmic}.
Recent advances in deep learning have enabled policy learning that directly predicts high-dimensional, continuous actions from image observations and robot states, with applications to diverse robotic manipulation tasks such as object manipulation, contact-rich tasks, and bimanual operation~\citep{vla_survey2025}.

However, most existing imitation learning tasks implicitly assume that the manipulation target and task goal can be determined from visual observations.
This assumption can break down under lighting variation, occlusion, the presence of multiple objects with identical or similar appearance, or when an object's internal state or events do not appear in images~\citep{playitbyear}.
To mitigate such partial observability, research integrating modalities such as touch and sound in addition to vision has been pursued~\citep{seehearfeel,vtt2022,sts_il2024,vitamin2024,tacthru2025,neuralfeels2024}.

In this work, we focus on acoustic information, which contains diverse cues that do not appear in object appearance.
Contact and collision sounds reflect object material and contact state~\citep{seehearfeel,thatsoundsright,hearingtouch2024,maniwav,multigen2025}, and sounds accompanying interactions such as pouring, shaking, and flowing provide cues for estimating the amount or flow rate of granular materials~\citep{clarke2018learning,wilson2019pouring} and liquid height~\citep{liang2020audiopouring,du2021robotpouring}.
Furthermore, through active exploration such as shaking, tapping, and rubbing objects, a robot can elicit acoustic cues unavailable from vision alone~\citep{eppe2018interactive,caver2025,mosaic2024}.
Importantly, acoustic information can carry not only signal information (what is happening) but also spatial information (where it is happening).

This property gives rise to task settings that differ from conventional vision-centric manipulation.
For example, when selecting a sound-emitting object among visually identical objects, determining object state or destination from timbre, or deciding the next action from acoustic cues obtained by shaking an object, the policy must integrate acoustic spatial and acoustic signal information, in addition to visual object layout, to generate actions.
We introduce such tasks, in which a spatially distributed sound environment directly governs target selection, goal determination, or active exploration, as \emph{acoustic-aware manipulation tasks}.

Existing acoustic robotic manipulation research has mainly used spectrograms from contact or single microphones to identify timbre, contact events, and object state~\citep{playitbyear,seehearfeel,maniwav,hearingtouch2024,audiovla}.
Robot audition, centered on sound source localization, tracking, and separation with microphone arrays, has targeted dialogue, navigation, and source tracking, but its integration into manipulation policies has been limited~\citep{nakadai2020robotaudition,robotaudition_review2025}.

We therefore propose the Spatial-Spectral Audio Action (S2A2) framework, which explicitly feeds the sound source position estimated from microphone arrays into an imitation learning policy as an acoustic spatial map.
Specifically, we combine an acoustic spatial map, a 2D map of the likelihood of the source position estimated by the MUSIC method~\citep{schmidt1986music,doa_survey2025}, with a spectrogram extracted by applying Spotforming~\citep{spotforming_nmf} to the estimated source position, and integrate them with visual information and proprioception.
The policy can thus use where a sound is produced from the acoustic spatial map and what kind of sound it is from the spectrogram.
Whereas existing imitation learning with sound feeds only spectrograms or raw waveforms to the policy and thus treats the source position only implicitly, our main contribution is to introduce the source position obtained from microphone arrays into imitation learning as an explicit spatial representation.

We apply S2A2 to the acoustic-aware manipulation tasks and evaluate it using a manipulator robot and multiple microphone arrays in both simulation and the real world, verifying that spatial acoustic representations function effectively in real manipulation.

Our contributions are threefold:
\begin{itemize}
	\item We propose S2A2, which explicitly feeds the source position estimated from microphone arrays into an imitation learning policy as an acoustic spatial map, together with a spectrogram extracted by Spotforming, vision, and proprioception. To our knowledge, this is the first work to explicitly introduce microphone-array-based source position information into a learned manipulation policy.
	\item We introduce acoustic-aware manipulation tasks, in which a spatially distributed sound environment is involved in target selection, goal determination, and active exploration, as diagnostic tasks that isolate the contributions of source localization, sound identification, and exploration.
	\item We integrate S2A2 into ACT, Diffusion Policy, VQ-BeT, and $\pi_0$ and systematically evaluate the contribution of acoustic information to policy learning in simulation, including a controlled comparison with a multi-channel raw-audio baseline, and further validate S2A2 through real robot ablations.
\end{itemize}

	\begin{figure}[t]
		\centering
		\includegraphics[keepaspectratio, width=14.5cm]{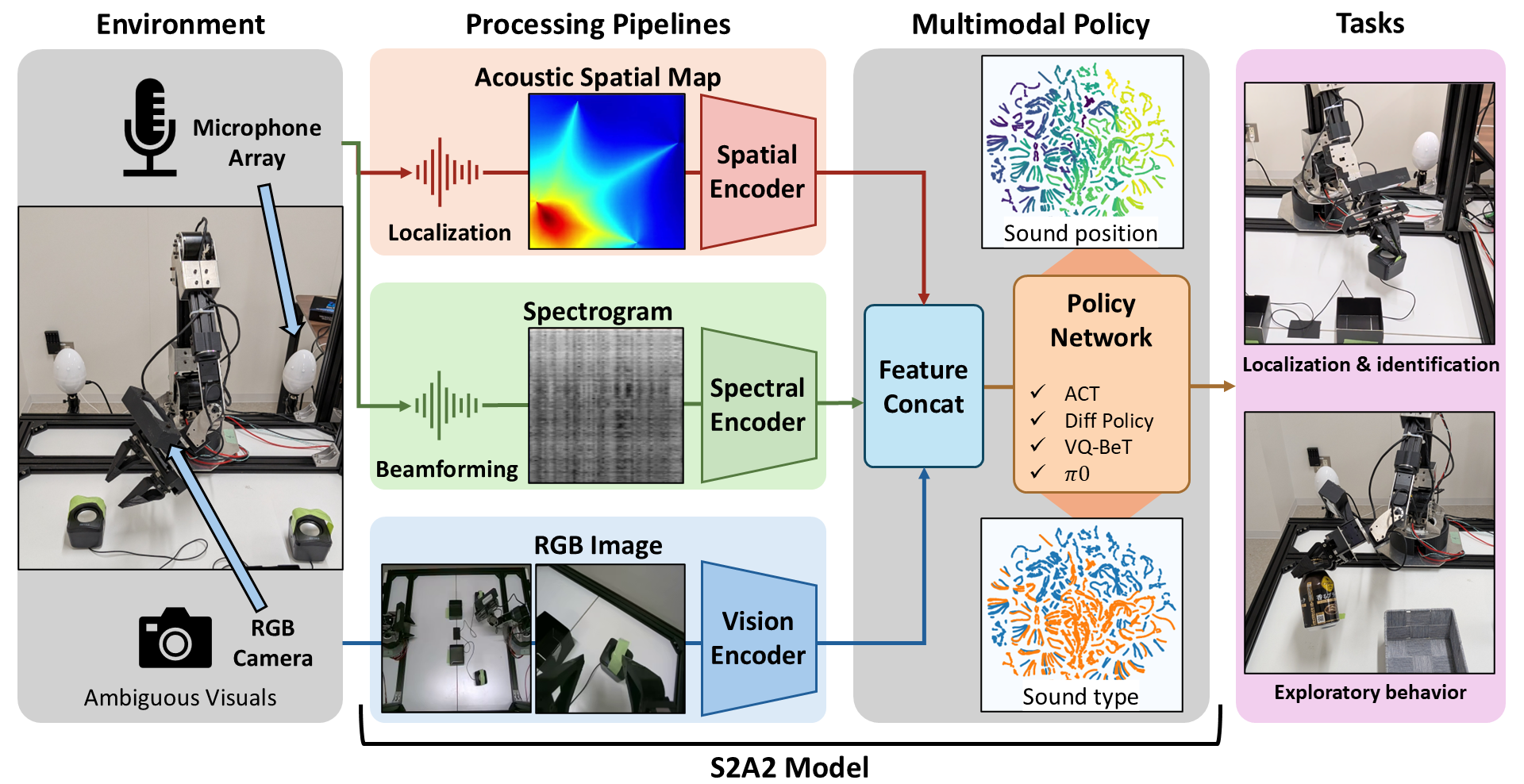}
		\caption{Overview of the S2A2 framework.
        }
		\label{fig:fig1}
	\end{figure}

\section{Related Work}

\subsection{Imitation Learning and Multimodal Manipulation}
Most recent imitation learning methods for robot manipulation rely primarily on visual observations~\citep{rmp_survey2025}. Vision-Language-Action models such as RT-2~\citep{brohan2023rt2} and OpenVLA~\citep{kim2024openvla} further connect vision- and language-based task specifications to action generation for general manipulation policies~\citep{vla_survey2025}. However, when manipulation-relevant information is not visible from appearance, or when objects are visually indistinguishable, vision and language alone cannot sufficiently resolve physical ambiguity.

To address this limitation, non-visual sensor modalities have been integrated into imitation learning. In particular, touch captures post-contact mechanical states and local object properties, and visuo-tactile policies have been shown to outperform vision-only policies in dense packing~\citep{seehearfeel}, contact-rich manipulation~\citep{vtt2022,sts_il2024,vitamin2024,tacthru2025}, and in-hand manipulation~\citep{neuralfeels2024}. The acoustic representations studied in this work are complementary to touch and language, suggesting the potential for broader multimodal manipulation when combined with these modalities.

\subsection{Manipulation with Acoustic Modality}
The acoustic modality has been used in many perception tasks, including material identification~\citep{eppe2018interactive,jin2019openenv}, estimation of granular material flow rate and amount~\citep{clarke2018learning,wilson2019pouring}, and liquid height estimation~\citep{liang2020audiopouring,du2021robotpouring}.
For robotic manipulation, policy learning~\citep{seehearfeel,maniwav} and action generation~\citep{thatsoundsright} using contact sounds have been proposed, and pretraining on large-scale acoustic datasets~\citep{hearingtouch2024} and integration with VLAs~\citep{audiovla} have been studied.

Beyond contact sounds, The Sound of Simulation~\citep{multigen2025} addresses sim-to-real policy learning with generative acoustic data augmentation, and CAVER~\citep{caver2025} and MOSAIC~\citep{mosaic2024} address the integration of active exploration and acoustic information.
Play it by Ear~\citep{playitbyear} combines vision and sound for manipulation under occlusion.
These studies mainly use sound from a single microphone near the gripper or from contact microphones attached to objects to identify contact events or object states. To our knowledge, no prior work has explicitly fed the source position estimated by microphone arrays into a learned manipulation policy.
In this work, in addition to the spectrogram, we feed the source position estimated from microphone arrays into the policy as an acoustic spatial map, which provides the policy with the information needed to associate a sound with a manipulation target in the image.

\subsection{Robot Audition and Microphone Array Processing}
Robot audition has studied sound source localization, tracking, and separation using microphone arrays in the presence of noise encountered in robot operating environments~\citep{nakadai2020robotaudition,robotaudition_review2025}.
Related work using multiple microphone arrays includes calibration with asynchronous arrays~\citep{wang2024slamcalib}, acoustic scene reproduction in virtual space~\cite{gabriel20192d}, and source tracking~\citep{yamada2020sound}.
In particular, Spotforming~\citep{spotforming_nmf}, which combines multiple arrays with NMF, emphasizes sources in a target region and suppresses ambient and co-directional noise.
This work differs from these robot audition studies in that the source position and the separated sound obtained by such microphone array processing are integrated with visual information and proprioception and used for imitation learning of manipulation.

\section{Spatial-Spectral Audio Action (S2A2) framework}
\label{sec:s2a2_framework}
Unlike existing imitation learning with sound, in which the policy has to learn the source position implicitly from spectrograms or raw waveforms, the S2A2 framework explicitly feeds the source position estimated from microphone arrays into the policy as an acoustic spatial map and integrates it with a spectrogram extracted by Spotforming, visual information, and proprioception to generate actions.

As shown in Fig.~\ref{fig:fig1}, the S2A2 framework consists of an environment setup, the S2A2 model, and the acoustic-aware manipulation tasks.
The S2A2 model consists of processing pipelines that extract per-modality features from observations and a multimodal policy that integrates them to output the next action.

\subsection{Problem Formulation}
\label{sec:problem}
The S2A2 framework targets acoustic-aware manipulation tasks in which the manipulation target or destination cannot be uniquely determined from vision alone, and determines actions using acoustic spatial and acoustic signal information in addition to vision.

We formulate this as an imitation learning problem conditioned on visual and auditory observations. Each observation $o_t$ consists of an RGB image $\mathbf{I}_t$, multi-channel acoustic information $A_t$, and proprioception $s_t$; from $A_t$ we obtain an acoustic spatial map $S_t$ and a spectrogram $G_t$.
Based on these observations, the S2A2 model learns a policy $\pi_\theta$ that predicts an action sequence $a_{t:t+H-1}$ up to $H$ steps ahead. Training minimizes the imitation learning loss measuring the deviation between predicted and expert actions over a dataset $\mathcal{D}$ of expert trajectories:
\begin{equation}
\min_{\theta}\ \mathbb{E}_{(o_t,a_{t:t+H-1})\sim\mathcal{D}}
\left[\mathcal{L}_{\mathrm{BC}}\left(\pi_\theta(o_t),a_{t:t+H-1}\right)\right].
\end{equation}
The specific loss depends on the policy network; details are given in Section~\ref{sec:policy} and Appendix~\ref{sec:appendix_architecture}.

\subsection{Acoustic-Aware Manipulation Tasks}
\label{sec:tasks}
To evaluate the role of acoustic information in manipulation decisions, we use four manipulation tasks, termed acoustic-aware manipulation tasks.
These four tasks are not intended to cover acoustic reasoning in real environments in general; they are diagnostic tasks designed to isolate, under controlled conditions, the contributions of source localization, sound identification, and exploration to manipulation.
Table~\ref{tab:tasks} summarizes them.
In every task, the environment contains target objects and destination boxes, and the manipulator must grasp the appropriate object and drop it into the specified box.
Object initial positions are randomized, while box positions are fixed.

The \textbf{Localization task} uses two visually indistinguishable objects, only one of which continuously emits sound; the manipulator grasps the sounding object and places it in the box.
A single sound is used across all episodes, so the information needed to select the grasp target is the spatial position of the source.

The \textbf{Identification task} uses one object and two boxes.
The object continuously emits one of two sounds, and the manipulator transports it to the box corresponding to that sound.
The box position for each sound is fixed across episodes.
Since there is a single sounding object, the grasp target is uniquely determined from vision and acoustic spatial information is unnecessary; sound identification is required to select the correct destination.

The \textbf{Localization \& Identification task} (L\&I Task) uses two visually indistinguishable objects and two boxes.
Only one object continuously emits sound, which is one of two sounds.
The manipulator identifies the grasp target from acoustic spatial information and transports it to the corresponding box based on the sound.
The box position for each sound is fixed across episodes.
Thus, this task requires source localization for target selection and sound identification for destination selection.

The \textbf{Exploratory task} uses two visually indistinguishable objects and one box.
Neither object emits sound at rest, and only one emits sound when moved.
The manipulator must grasp and shake each object in turn to check for sound and place the sounding object in the box.
This exploratory behavior is acquired by imitating demonstrations in which the object is lifted and shaken to check for sound; S2A2 does not discover information-gathering actions on its own.

\begin{wraptable}{r}{0.48\textwidth}
  \centering
  \setlength{\tabcolsep}{0pt}
  \caption{Acoustic capabilities relevant to each task in the acoustic-aware manipulation tasks.}
  \label{tab:tasks}
  \begin{tabular}{p{0.15\textwidth} p{0.09\textwidth}
    >{\centering\arraybackslash}p{0.09\textwidth}
    >{\centering\arraybackslash}p{0.09\textwidth}}
    \hline
    Task & Space & Signal & Explore \\
    \hline
    Localization & $\checkmark$ & -- & -- \\
    Identification & -- & $\checkmark$ & -- \\
    L\&I & $\checkmark$ & $\checkmark$ & -- \\
    Exploratory & $\checkmark$ & $\checkmark$ & $\checkmark$ \\
    \hline
  \end{tabular}
\end{wraptable}

\subsection{Processing Pipeline}
\label{sec:pipelines}
The S2A2 model has three pipelines that process acoustic spatial information, acoustic signal information, and visual information.
Each pipeline consists of preprocessing and an encoder appropriate to its input modality.
Except for the RGB image pipeline, which directly uses the vision encoder of the underlying policy, this section describes the acoustic spatial map pipeline and the spectrogram pipeline.

\subsubsection{Acoustic Spatial Map Pipeline}
\label{sec:doa}
As shown in Fig.~\ref{fig:fig2}, all microphone arrays are placed on the same plane as the workspace.
The acoustic spatial map pipeline estimates the likelihood of source direction from the multi-channel acoustic information recorded by the arrays and projects it onto a 2D map corresponding to the workspace.
Specifically, the multiple signal classification (MUSIC) method~\citep{schmidt1986music} is applied to the multi-channel signal of each array.
Input audio is acquired at a 16~kHz sampling rate and transformed to the frequency domain by a short-time Fourier transform (STFT) with an FFT length of 512 and 50\% overlap.

The log-normalized spatial spectrum per direction is used as the direction-of-arrival (DOA) score. With an attenuation term based on distance from the array center, it is projected onto a $224\times224$ 2D plane corresponding to the array plane of the workspace, yielding the acoustic spatial map $S\in\mathbb{R}^{224\times224\times N}$, where $N$ is the number of arrays. Details are in Appendix~\ref{sec:app_doa}.
The acoustic spatial map $S$ is fed to a spatial encoder to produce features usable by the policy network.
The spatial encoder is a ResNet-18 with the input channel count extended to $N$, trained from scratch, and outputs a feature map.

\subsubsection{Spectrogram Pipeline}
\label{sec:spec}
The spectrogram pipeline extracts the acoustic signal information of the target source.
From the acoustic spatial map $S$, the source presence likelihood is computed, and peak extraction selects source candidates.
Since only one source is present at a time, the highest peak is taken as the candidate.
Source separation by Spotforming~\citep{spotforming_nmf} is applied to the candidate to generate the spectrogram. Details are in Appendix~\ref{sec:app_doa}.
The spectrogram is fed to a spectral encoder, a ResNet-18 with a single input channel trained from scratch, to produce features usable by the policy network.

\subsection{Multimodal Policy}
\label{sec:policy}
The multimodal policy integrates the per-modality features from the processing pipelines with proprioception to generate robot actions.
We adopt four policy networks for action generation: ACT~\citep{act}, Diffusion Policy~\citep{diffusionpolicy}, VQ-BeT~\citep{vqbet}, and $\pi_0$~\citep{pi0}.

As described in Section~\ref{sec:pipelines}, each pipeline outputs a feature map.
The format for feeding these feature maps to the policy network differs by policy architecture; details are in Appendix~\ref{sec:appendix_architecture}.

\section{Experiments}
\label{sec:result}

\subsection{Simulation Experiments}
\label{sec:sim_results}
\begin{figure}[t]
		\centering
		\includegraphics[keepaspectratio, width=14cm]{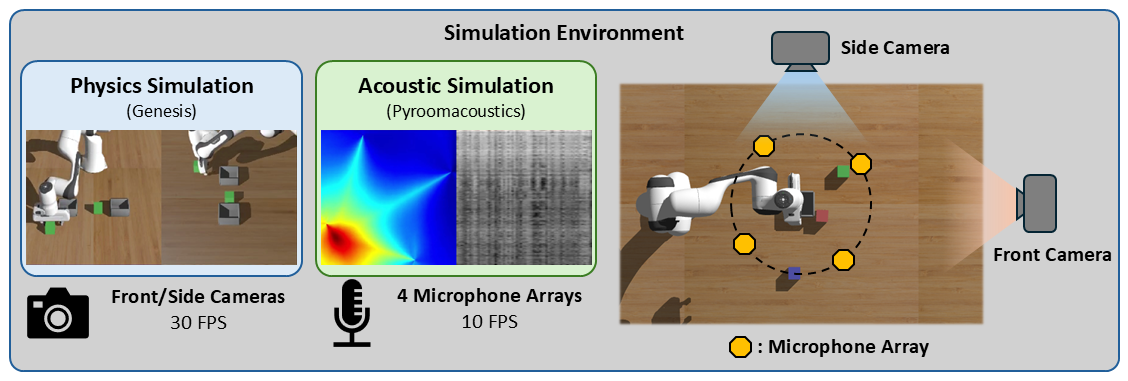}
		\caption{Simulation settings. A simulation environment integrating the physics simulator Genesis and the acoustic simulator Pyroomacoustics. RGB images are updated at 30~FPS and acoustic information from the microphone arrays at 10~FPS. The workspace center is 500~mm in front of the Franka Emika Panda, with four microphone arrays placed at equal intervals on a circle of radius 300~mm around it. Two cameras (side and front) are set to view the workspace center.}
		\label{fig:fig2}
	\end{figure}

We first conduct simulation experiments to examine the relationship between the processing pipelines in the S2A2 framework and the acoustic-aware manipulation tasks defined in Section~\ref{sec:tasks}.
We compare four conditions by varying the combination of processing pipelines:
(\textbf{Baseline}) uses only the RGB image pipeline, with no acoustic information;
(\textbf{S2A2}) the standard proposed method using all three pipelines;
(\textbf{w/o Spec.}) removes the spectrogram pipeline, using visual information and acoustic spatial information from the acoustic spatial map;
(\textbf{w/o Spat.}) removes the acoustic spatial map pipeline, using visual information and acoustic signal information from the spectrogram.
In addition, to examine the effect of these explicit acoustic representations, we compare, for ACT and Diffusion Policy, a condition (\textbf{Raw}) that instead feeds the multi-channel raw waveforms of the four microphone arrays to the policy.
The Raw encoder takes the same 1-s audio window as the spectrogram and extracts features from the waveforms with a wav2vec-style 1D convolutional neural network trained from scratch; the number of acoustic tokens for ACT and the dimension of the acoustic feature vector for Diffusion Policy are matched to those of S2A2 (details in Appendix~\ref{sec:appendix_raw}).
In all conditions, the multimodal policy structure is identical, concatenating the features output by each pipeline and feeding them to the policy network.
As shown in Fig.~\ref{fig:fig2}, we build a simulation environment integrating the physics simulator Genesis~\citep{Genesis} and the acoustic simulator Pyroomacoustics~\citep{pyroomacoustics} (details in Appendix~\ref{sec:app_simu}).
Other optimization settings and internal model parameters are summarized in Appendix~\ref{sec:appendixA}. Each task is trained with three random seeds, and the evaluation metric is the fraction of 100 trials in which the correct object is grasped and placed in the correct box within about 23 seconds (700 steps).
We use 100 demonstration episodes for the localization and identification tasks and 200 for the L\&I and exploratory tasks.

\begin{figure}[t]
		\centering
		\includegraphics[keepaspectratio, width=14cm]{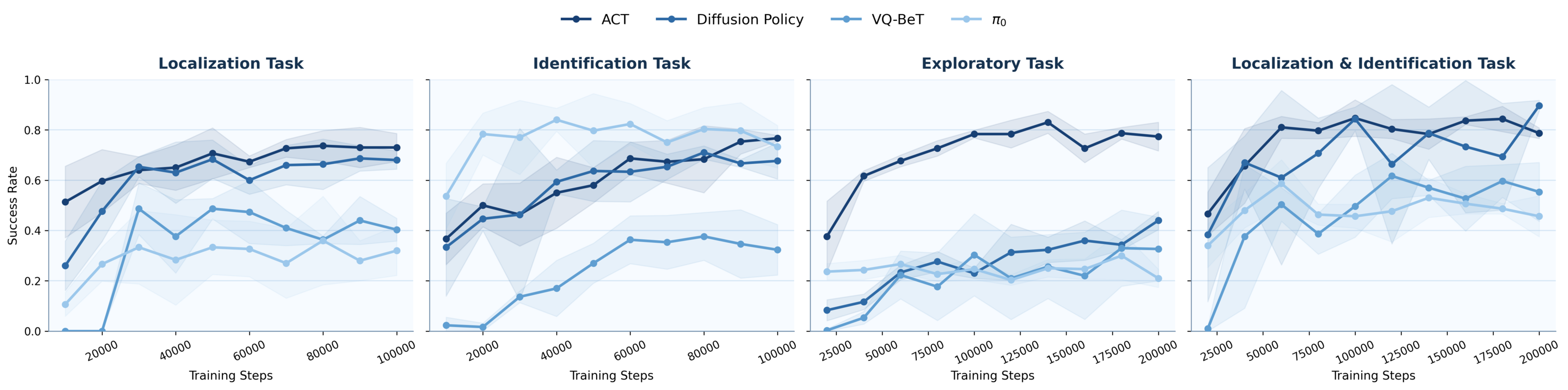}
		\caption{Task success rate versus training steps. Training curves of the four policies (ACT, Diffusion Policy, VQ-BeT, $\pi_0$) of the S2A2 model on the four tasks. The localization and identification tasks were evaluated every 10,000 steps up to 100,000 steps; the L\&I and exploratory tasks every 20,000 steps up to 200,000 steps.}
		\label{fig:fig3}
\end{figure}

	\begin{table*}[t]
		\centering
		\small
		\caption{Comparison of task success rates (\%) across all tasks (Mean $\pm$ STD). Raw denotes the baseline that takes multi-channel raw waveforms as input (ACT and Diffusion Policy only). The best value within each policy is shown in bold.}
		\label{tab:results}
		\begin{tabular}{llcccc}
			\hline
			Model & Policy & localization task & identification task & L\&I Task & exploratory task \\
			\hline
			Baseline & ACT & 20.3 $\pm$ 7.2 & 39.3 $\pm$ 4.0 & 11.0 $\pm$ 3.5 & 35.0 $\pm$ 5.3 \\
			 & Diffusion Policy & 13.3 $\pm$ 1.2 & 41.3 $\pm$ 2.5 & 9.3 $\pm$ 6.1 & 26.3 $\pm$ 0.6 \\
			 & VQ-BeT & 12.7 $\pm$ 6.8 & 24.0 $\pm$ 6.9 & 7.7 $\pm$ 1.2 & 17.3 $\pm$ 5.8 \\
			 & $\pi_0$ & 18.3 $\pm$ 2.3 & 32.3 $\pm$ 2.1 & 7.0 $\pm$ 4.0 & 9.0 $\pm$ 1.0 \\
			\hline
			S2A2 & ACT & 73.0 $\pm$ 5.6 & 76.7 $\pm$ 1.2 & \textbf{78.7 $\pm$ 2.1} & \textbf{77.3 $\pm$ 5.7} \\
			 & Diffusion Policy & 68.0 $\pm$ 3.6 & 67.7 $\pm$ 7.2 & \textbf{89.7 $\pm$ 2.1} & 44.0 $\pm$ 3.6 \\
			 & VQ-BeT & 40.3 $\pm$ 4.5 & 32.3 $\pm$ 10.0 & \textbf{55.3 $\pm$ 11.7} & \textbf{32.7 $\pm$ 12.7} \\
			 & $\pi_0$ & \textbf{32.0 $\pm$ 9.9} & 73.3 $\pm$ 8.4 & \textbf{45.7 $\pm$ 8.1} & 18.7 $\pm$ 4.2 \\
			\hline
			w/o Spec. & ACT & \textbf{74.7 $\pm$ 4.7} & 44.3 $\pm$ 2.3 & 45.0 $\pm$ 3.6 & 75.3 $\pm$ 5.0 \\
			 & Diffusion Policy & \textbf{68.3 $\pm$ 13.7} & 32.7 $\pm$ 3.1 & 46.0 $\pm$ 5.3 & 37.3 $\pm$ 8.1 \\
			 & VQ-BeT & \textbf{44.7 $\pm$ 5.5} & 32.7 $\pm$ 3.1 & 32.3 $\pm$ 7.5 & 20.0 $\pm$ 20.0 \\
			 & $\pi_0$ & 17.3 $\pm$ 5.0 & 40.0 $\pm$ 11.4 & 14.3 $\pm$ 2.9 & \textbf{26.7 $\pm$ 3.2} \\
			\hline
			w/o Spat. & ACT & 15.3 $\pm$ 4.0 & 74.3 $\pm$ 7.0 & 25.7 $\pm$ 3.1 & 72.3 $\pm$ 8.4 \\
			 & Diffusion Policy & 15.3 $\pm$ 1.2 & \textbf{80.3 $\pm$ 3.8} & 8.7 $\pm$ 3.8 & 6.3 $\pm$ 2.3 \\
			 & VQ-BeT & 12.3 $\pm$ 3.8 & \textbf{49.3 $\pm$ 9.8} & 19.0 $\pm$ 1.0 & 25.0 $\pm$ 3.6 \\
			 & $\pi_0$ & 14.7 $\pm$ 5.7 & \textbf{78.7 $\pm$ 6.0} & 19.7 $\pm$ 3.8 & 9.3 $\pm$ 10.4 \\
			\hline
			Raw & ACT & 43.3 $\pm$ 2.5 & \textbf{80.7 $\pm$ 4.2} & 48.7 $\pm$ 6.8 & 65.0 $\pm$ 8.2 \\
			 & Diffusion Policy & 20.3 $\pm$ 8.0 & 68.7 $\pm$ 4.0 & 24.7 $\pm$ 4.2 & \textbf{50.7 $\pm$ 1.5} \\
		\end{tabular}
	\end{table*}

\subsubsection{Task Success Rate Comparison}

Table~\ref{tab:results} shows success rates across the four tasks, and Fig.~\ref{fig:fig3} shows their progression over training steps.
The results generally follow the acoustic information required by each task.
In the localization task, conditions that include the acoustic spatial map perform substantially better than conditions without it.
In the identification task, conditions that provide acoustic signal information perform well, while the acoustic spatial map is not necessary.
In the L\&I task, which requires both source localization and sound identification, S2A2 achieves the highest success rate for all four policy architectures.
The exploratory task shows a stronger dependence on the policy architecture, as discussed in Appendix~\ref{sec:results}.

Adding an acoustic processing pipeline that is not required by a task can reduce performance under the limited amount of demonstration data used here.
Table~\ref{tab:results} also compares S2A2 with the multi-channel Raw baseline for ACT and Diffusion Policy.
In the localization and L\&I tasks, where the source position is required for selecting the manipulation target, S2A2 substantially outperforms Raw for both policies.
In the identification and exploratory tasks, Raw remains competitive.
These results suggest that, under the amount of demonstration data used in this study, explicitly providing the source position is particularly useful when the action depends on where the sound is produced.

\subsubsection{Latent Feature Visualization}
\label{sec:visualization}
\begin{figure}[t]
		\centering
		\includegraphics[keepaspectratio, width=14cm]{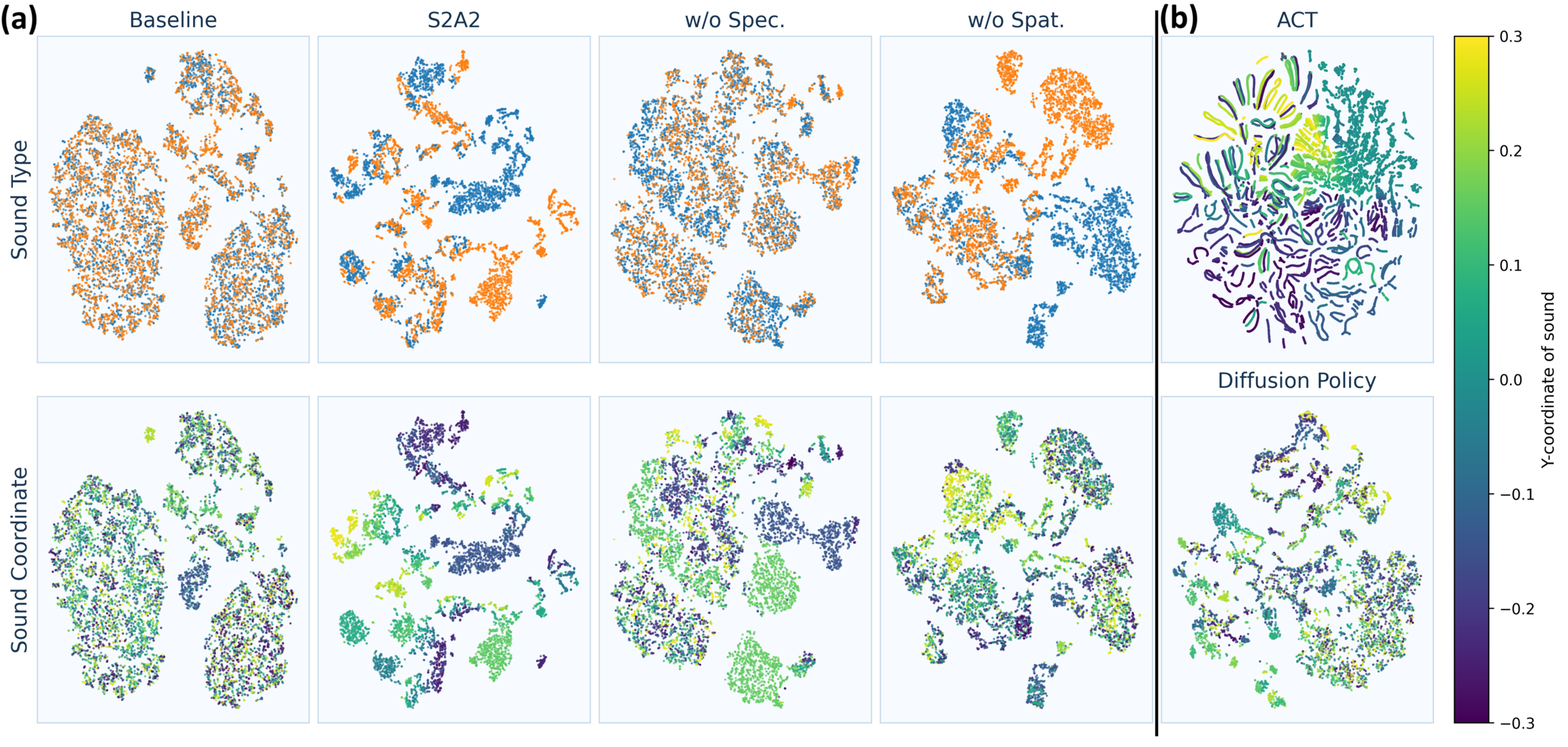}
		\caption{Latent space visualization. (a) Comparison for Diffusion Policy on the L\&I task. The top row is colored by the sound type, and the bottom row encodes the Y coordinate of the sounding object as a colormap. In the top row, sound A is orange and sound B is blue. (b) Intermediate-layer latent representations of ACT and Diffusion Policy on the exploratory task, separated by t-SNE, with the Y coordinate of the sounding grasp target encoded as a colormap. Results for all policies are in Appendix~\ref{sec:appendixB}.}
		\label{fig:fig4}
	\end{figure}

We visualize the intermediate-layer latent features of ACT and Diffusion Policy in Fig.~\ref{fig:fig4}, projected to 2D by t-SNE~\citep{vandermaaten2008tsne}, and discuss their relation to task success rate.
The visualization method and results for all policies are summarized in Appendix~\ref{sec:appendixB}.

Fig.~\ref{fig:fig4}(a) shows the latent space of Diffusion Policy on the L\&I task.
The relatively high success rate of the S2A2 model is also reflected in the latent space.
In the S2A2 model, both the sound type and the sound coordinate (the Y coordinate of the target position) are relatively well separated and structured, whereas in models lacking the corresponding information the separation is weaker.
Both quantitative and qualitative evaluations therefore support that acoustic spatial and acoustic signal information play independent roles in the latent space.

Fig.~\ref{fig:fig4}(b) shows the latent space of Diffusion Policy and ACT on the exploratory task.
ACT's latent space forms a clear structure corresponding to source position, whereas Diffusion Policy's is less structured.
This difference in latent structuring can be interpreted as the source of the performance difference on the exploratory task.

\subsection{Real Robot Experiments}
\label{sec:real}
\begin{wraptable}{r}{0.5\textwidth}
  \centering
  \footnotesize
  \setlength{\tabcolsep}{3pt}
  \caption{Real robot success rates (\%) over 100 trials. Best in bold.}
  \label{tab:real}
  \begin{tabular}{lcccc}
    \hline
    Task & Baseline & w/o Spec. & w/o Spat. & S2A2 \\
    \hline
    L\&I & 5 & 18 & 3 & \textbf{72} \\
    Exploratory & 12 & 33 & 27 & \textbf{50} \\
    \hline
  \end{tabular}
\end{wraptable}
To verify the simulation findings in the real world, we use only the right arm of ILOHA, an open-source ALOHA-type~\citep{act} bimanual manipulator.
The six joints and the 1-DoF gripper are treated as state and action, so both are 7-dimensional.
Visual information is obtained from a top-down view and a RealSense camera mounted on the end-effector, cropped and resized to $224\times224$ RGB images.
Acoustic information is acquired with four circular microphone arrays (TAMAGO-03), each with $8$ microphones on a circle of radius $35$~mm, at a $16$~kHz sampling rate.
The arrays are placed at the vertices of a rectangle ($0.6$~m by $1.2$~m, top-down) surrounding the workspace, with their centers approximately $0.14$~m above the working plane.

As the policy, we use ACT, which performed stably on both tasks in simulation, and compare the same four conditions as in simulation: the RGB-only baseline, w/o Spec., w/o Spat., and S2A2.
For each task, 100 demonstration episodes were collected by human teleoperation and shared across the four conditions (Appendix~\ref{sec:app_real}).
We select the L\&I task, which requires both localization and identification, and the exploratory task, which requires active exploration; Table~\ref{tab:real} shows the success rates over 100 trials.
In both tasks, S2A2 achieves the highest success rate.
In the L\&I task, S2A2 achieves a success rate of 72\%, compared with 5\% for the RGB-only baseline, 18\% for w/o Spec., and 3\% for w/o Spat., showing that combining the acoustic spatial map and the spectrogram yields substantially higher performance than using either acoustic representation alone.
In the exploratory task, both single-representation conditions improve over the RGB-only baseline, while S2A2 achieves the highest success rate.
This suggests that either acoustic representation can provide useful information for selecting actions after acoustic observation, while combining both representations provides the highest performance in this setting.

	\begin{figure}[t]
		\centering
        \includegraphics[keepaspectratio, width=10cm]{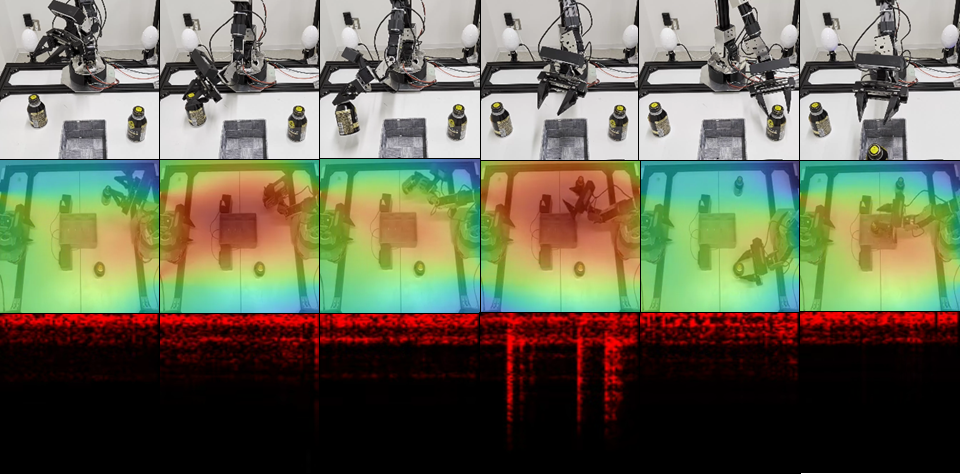}
		\caption{Example real robot sequence of the S2A2 model with ACT on the exploratory task. Top: the robot. Middle: the acoustic spatial map summed over channels, shown as a colormap overlaid on the top-down camera image. Bottom: the visualized spectrogram.}
		\label{fig:fig5}
	\end{figure}

\section{Limitations}
Our work has several limitations.
First, the tasks are diagnostic tasks that isolate source localization, sound identification, and exploration; more complex acoustic cues such as material, hollowness, and fill level, and more complex manipulation, are not evaluated.
Real robot validation was conducted on a single robot in a limited environment, and generalization to diverse platforms including bimanual and mobile robots, to diverse environmental conditions, and especially to environments with multiple simultaneous sound sources is left for future work.
Second, the acoustic spatial map assumes a known array placement; in our setup, a localization error of 5 degrees corresponds to at most about 5 cm on the workspace, so no precise calibration was needed, but calibration becomes more important in a larger workspace.
Handling long missing or delayed acoustic observations depends on the temporal modeling of the underlying policy rather than on S2A2 itself.
Third, the exploratory behavior is imitated from demonstrations; acquiring information-gathering behavior beyond them would require mechanisms such as reinforcement learning.
The efficiency of using acoustic information also depends on the policy architecture, with limited improvement under some conditions for $\pi_0$ and VQ-BeT; designing effective integration of acoustic information into large-scale pretrained models and incorporating sim-to-real methods~\citep{multigen2025} are important future directions.

\section{Conclusion}
\label{sec:conclusion}
We proposed S2A2, an imitation learning framework that explicitly feeds the source position estimated from microphone arrays into the policy as an acoustic spatial map together with a Spotforming spectrogram and vision, and introduced acoustic-aware manipulation tasks for its evaluation.
Simulation experiments showed the importance of representations matched to the acoustic capabilities a task requires, that unnecessary acoustic modalities can degrade performance, that the efficiency of using acoustic information depends on the policy architecture, and that in tasks requiring the source position S2A2 substantially outperforms a multi-channel raw-audio baseline.
Real robot experiments further confirmed that S2A2 outperforms both the vision-only baseline and the conditions using only one acoustic representation.

\clearpage
\acknowledgments{This work was supported by JST CREST(No. JPMJCR22D3), JST Moonshot R\&D (JPMJMS2024), JSPS KAKENHI (JP26H02532, JP26H02525), and RIKEN TRIP initiative (AGIS).}

\bibliography{main}

\clearpage

\appendix


\section{Implementation Details} 
\label{sec:appendixA}
This appendix describes the implementation details of the S2A2 model.
Specifically, Section~\ref{sec:appendix_architecture} provides the details of the policy network architectures and hyperparameters, Section~\ref{sec:appendix_audio_params} provides the details of the audio processing, and Section~\ref{sec:appendix_sim_real} describes the changes introduced to the S2A2 model configuration for the real robot experiments relative to simulation.

\subsection{Policy Architecture and Hyperparameters} 
\label{sec:appendix_architecture}
Each policy architecture and each input condition is trained independently.
The encoders for the acoustic spatial map and the spectrogram do not use pretrained weights and are trained from random initialization; the initialization of the RGB encoder and of the pretrained parts of $\pi_0$ follows the description of each policy below.
This section describes, for the four policy networks evaluated in this work, the feature extraction from observations, the action generation, and the training and inference procedures.
S2A2 denotes the standard model that uses image information, the acoustic spatial map, the spectrogram, and proprioception. w/o Spat.\ denotes the model without the acoustic spatial map pipeline, and w/o Spec.\ denotes the model without the spectrogram pipeline.

\subsubsection{ACT}
The ACT vision encoder uses ResNet-18, initialized with weights pretrained on ImageNet.
The encoders for the acoustic spatial map and the spectrogram use the ResNet-18 architecture with only the number of input channels modified, and are trained from scratch without pretraining.
The feature maps output by the RGB image, acoustic spatial map, and spectrogram encoders are mapped to a sequence of $512$-dimensional tokens by a $1\times1$ convolution.
After adding positional embeddings to these tokens and appending the proprioception as a state token, the tokens are fed into a transformer composed of an encoder and a decoder.
Within ACT, a VAE maps action sequences into a latent space. In this configuration, the transformer has a hidden dimension of $512$, $8$ attention heads, $4$ encoder layers, and $1$ decoder layer, and the VAE latent dimension is $32$.
The decoder outputs an action chunk of length $H_{\mathrm{act}}$.
During training, the VAE is used jointly and the loss function in Eq.~(\ref{eq:eq_j}) is minimized.
\begin{equation}
\label{eq:eq_j}
\mathcal{L}_{\mathrm{ACT}} = \frac{1}{H_{\mathrm{act}} d_a}\sum_{h=0}^{H_{\mathrm{act}}-1} m_h\left\|\hat{a}_{t+h}-a_{t+h}\right\|_1 + \beta_{\mathrm{KL}}\, D_{\mathrm{KL}}\left(q_\phi(z\mid a_{t:t+H_{\mathrm{act}}-1},s_t)\,\|\,\mathcal{N}(0,\mathbf{I})\right)
\end{equation}
Here, $d_a$ is the action dimensionality, $h$ is the time index within the action chunk, and $\hat{a}_{t+h}$ and $a_{t+h}$ are the predicted action and the expert action, respectively.
$m_h\in\{0,1\}$ is a mask that excludes the padding at the sequence ends from the loss computation, $s_t$ is the proprioception, $z$ is the VAE latent variable, $\phi$ denotes the parameters of the VAE encoder, and $q_\phi$ denotes the resulting latent distribution.
$\mathcal{N}(0,\mathbf{I})$ is the standard normal distribution with zero mean and identity covariance, $D_{\mathrm{KL}}$ is the Kullback--Leibler divergence, and the weight of this term is set to $\beta_{\mathrm{KL}}=10$.

\subsubsection{Diffusion Policy}
In Diffusion Policy, the vision encoder that processes the visual information is a Residual Network with 18 layers (ResNet-18). No pretrained weights are used; all parameters are randomly initialized.
Input images are resized to $84\times84$ pixels before being fed to the encoder.
The encoders that process the acoustic spatial map and the spectrogram (a time--frequency representation) used as acoustic information are ResNet-18 with only the number of input channels modified, and likewise use no pretrained weights.
The feature map output by each encoder is linearly projected onto a two-dimensional feature map with $n_{key}=32$ keypoint features.
A spatial softmax is then applied to each channel.
For each channel, spatial softmax normalizes the feature values across spatial locations to form a probability distribution over those locations.
Based on the resulting weights, a weighted average of the normalized image coordinates is computed, extracting the keypoint feature (the representative location of the feature) corresponding to each channel.
The extracted keypoint features are flattened into a one-dimensional vector and concatenated with the proprioception, which represents the robot's internal state such as joint angles and joint velocities. This vector is used as the global conditioning vector provided to the U-Net.
The channel counts of the U-Net used as the noise prediction model in Diffusion Policy are $(512, 1024, 2048)$.
During training, for an action sequence $x_0=a_{t-1:t+14}$ of length $H_d=16$, at a step $k$ sampled uniformly from the 100 diffusion steps of the Denoising Diffusion Probabilistic Model (DDPM), the noised $x_k$ obtained by adding noise $\epsilon\sim\mathcal{N}(0,\mathbf{I})$ is generated according to Eq.~(\ref{eq:eq_k}).
\begin{equation}
\label{eq:eq_k}
x_k=\sqrt{\bar{\alpha}_k}x_0+\sqrt{1-\bar{\alpha}_k}\epsilon
\end{equation}
Here, $\bar{\alpha}_k$ is a coefficient determined by the DDPM noise schedule. In this work, we use \texttt{squaredcos\_cap\_v2} as the noise scheduler.
Let the U-Net model be $\epsilon_\theta$ and let the vector encoding the observation conditioning be $g(o)$. We adopt \texttt{epsilon} as the prediction type and train the U-Net to directly estimate the noise $\epsilon$ added during the diffusion process, using the following loss function.
\begin{equation}
\label{eq:eq_l}
\mathcal{L}_{\mathrm{Diff}} =\mathbb{E}_{x_0,\epsilon,k} \left[\frac{1}{H_d d_a}\left\|\epsilon_\theta(x_k,k,g(o))-\epsilon\right\|_2^2\right]
\end{equation}
Here, $\theta$ denotes the learnable parameters of the U-Net and $o$ denotes the observation.
At inference time, 100 reverse-diffusion steps are performed, and 8 steps of actions starting from the last time step of the observation history are output as execution candidates.
The samples are clipped to $[-1, 1]$.

\subsubsection{VQ-BeT}
In VQ-BeT, the encoder that processes the visual information uses the same configuration as in Diffusion Policy: input images are resized to $84\times84$ pixels before being fed to the encoder, and a spatial softmax is applied to the feature map output by each encoder to extract keypoint features, as in Diffusion Policy.
The extracted features from each pipeline, the proprioception representing the robot's internal state such as joint angles and joint velocities, and the Action Query Tokens are fed to the policy network.
An Action Query Token is a learnable query vector for predicting future action sequences, with each token responsible for generating one action chunk.
The policy network is a decoder-only transformer (GPT) with a hidden dimension of $512$, $8$ attention heads, and $8$ transformer layers.
We represent the action output using a Residual Quantized Variational Autoencoder (RQ-VAE)~\citep{rqvae}. RQ-VAE quantizes a continuous-valued action sequence into multiple stages of discrete codes. In this work, an action chunk of length 5 is converted into a two-level sequence of discrete codes. The codebook (the dictionary of discrete representations) at each level has a size of 16. The RQ-VAE is trained for 20,000 steps independently of the policy network, after which the obtained codebook is fixed and used.
During training, the policy network is trained against the fixed RQ-VAE. Each Action Query Token predicts the probability distribution over the discrete codes at each level and a continuous offset for correcting the quantized action representation.
The discrete codes are intended to represent the coarse action of the action chunk, while the continuous offset corrects the quantization error.
Let the predicted action chunk be $\hat{A}_j$, the ground-truth action chunk be $A_j$, the ground-truth codes at the first and second levels be $c_j^{(1)}$ and $c_j^{(2)}$, respectively, and the corresponding predicted code distributions be $p_{\theta,j}^{(1)}$ and $p_{\theta,j}^{(2)}$. The loss function is then defined by the following equation.
\begin{equation}
\label{eq:eq_m}
\mathcal{L}_{\mathrm{VQ}} =\frac{1}{J}\sum_{j=1}^{J}\left[ \lambda_{\mathrm{off}}\frac{1}{H_q d_a}\|\hat{A}_j-A_j\|_1 +\lambda_1\,\mathrm{FL}(p_{\theta,j}^{(1)},c_j^{(1)}) +\lambda_2\,\mathrm{FL}(p_{\theta,j}^{(2)},c_j^{(2)}) \right]
\end{equation}
Here, $\theta$ denotes the learnable parameters of the policy network, $j$ is the index of the Action Query Token, $J=3$ is the total number of Action Query Tokens, $H_q=5$ is the action chunk length handled by one token, and $d_a$ is the dimensionality of the action space.
$\lambda_{\mathrm{off}}$, $\lambda_1$, and $\lambda_2$ are weighting coefficients that balance the contributions of the reconstruction loss for the continuous offset, the first-level discrete code prediction loss, and the second-level discrete code prediction loss, respectively. In this work, we set
$\lambda_{\mathrm{off}}=10000$,
$\lambda_1=5.0$, and
$\lambda_2=0.5$.
In addition, $\mathrm{FL}$ denotes the Focal Loss, a loss function that suppresses the contribution of easily classified samples and places greater emphasis on samples prone to misclassification. Letting $p_c$ be the predicted probability corresponding to the ground-truth code $c$, it is defined by the following equation.
\begin{equation}
\label{eq:eq_n}
\mathrm{FL}(p,c)
=
-(1-p_c)^2 \log p_c
\end{equation}
At inference time, the discrete code at each level is sampled from the predicted probability distribution, with a code sampling temperature of 0.1. The temperature controls the sharpness of the probability distribution, with smaller values making high-probability codes more likely to be selected. The sampled discrete codes are decoded into continuous values by the RQ-VAE decoder, and the predicted continuous offset is added to produce the final action output.

\subsubsection{$\pi_0$}
The original $\pi_0$ is a policy network that takes images and language as its primary inputs. In this work, we use SigLIP (Sigmoid Loss for Language--Image Pre-training) as the vision encoder that processes images.
In addition, the encoders for the acoustic spatial map and the spectrogram that process the acoustic information are ResNet-18 with a modified number of input channels.
These encoders use no pretrained weights and are randomly initialized.
The feature map output by each acoustic encoder is projected onto the hidden dimension of PaliGemma by a $1\times1$ convolution and then converted into a token sequence. The resulting acoustic tokens are concatenated with the output tokens of the vision encoder and placed on the prefix side of the input text token sequence. The proprioception representing the robot's internal state such as joint angles and joint velocities, the noised action sequence, and the timestep embedding are placed on the suffix side.
The model architecture combines PaliGemma (Gemma-2B) with an Action Expert (Gemma-300M) for action generation, and the model weights are initialized from \texttt{lerobot/pi0\_base}.
The state and action vectors are zero-padded to 32 dimensions. This unifies the input format with the fixed-dimensional representation expected by PaliGemma and the Action Expert.
In this method, the model is trained based on Flow Matching.
Flow Matching is a generative model that directly learns a velocity field along a continuous path between the data distribution and a simple noise distribution.
Instead of learning a sequential denoising process as in diffusion models, it is trained to predict the ideal velocity vector at each time.
Let the ground-truth action sequence be $a$, the standard normal noise of the same shape be $\epsilon$, and the Flow Matching time be $r\in[0,1]$. The target velocity $u_r$ used as the training objective is defined by the following equation.
\begin{equation}
\label{eq:eq_o}
x_r=r\epsilon+(1-r)a,\qquad u_r=\epsilon-a
\end{equation}
Here, $x_r$ is the intermediate state obtained by linearly interpolating between the ground-truth action sequence and the noise, and $u_r$ is the ideal velocity vector pointing from that state toward the noise distribution.
Following $\pi_0$~\citep{pi0}, to prevent $r$ from taking exactly $0$ or $1$, we add a small offset and use $r = 0.999b + 0.001$, where $b$ is sampled from the Beta distribution $\mathrm{Beta}(1.5,1.0)$.
Letting the velocity-field model be $v_\theta$, we minimize the following loss function so that the model predicts the target velocity $u_r$.
\begin{equation}
\label{eq:eq_p}
\mathcal{L}_{\pi_0} =\mathbb{E}_{a,\epsilon,r} \left[\frac{1}{H_\pi d_a}\left\|v_\theta(x_r,r,o)-u_r\right\|_2^2\right]
\end{equation}
Here, $\theta$ denotes the learnable parameters of the velocity-field model, $H_\pi=50$ is the action chunk length, $x_r$ is the noised action, $u_r$ is the target velocity from the ground-truth action toward the noise distribution, and $o$ is the observation conditioning, which includes the prefix and the proprioception.
At inference time, the action sequence is generated by initializing the trajectory from standard normal noise and numerically integrating the ordinary differential equation (ODE) defined by Flow Matching backward from $r=1$ to $r=0$ using the learned velocity field.
In this work, the ODE integration is performed in 10 steps, and the resulting action sequence is output as the execution candidate.
During training, gradient checkpointing is enabled to reduce memory usage, and gradient clipping is applied with a maximum gradient norm of $1.0$ to stabilize training.

\subsubsection{Raw-Audio Baseline}
\label{sec:appendix_raw}
To evaluate whether the explicit acoustic representations used in S2A2 are beneficial beyond simply providing multi-channel audio to the policy, we introduce a raw-audio baseline for ACT and Diffusion Policy.
This baseline replaces the acoustic spatial-map and spectrogram encoders with a single encoder that directly processes the multi-channel waveform, while keeping the subsequent policy-specific multimodal fusion as close as possible to that of S2A2.

\paragraph{Raw-audio observation}


The input consists of signals from four 8-channel microphone arrays, resulting in a 32-channel waveform sampled at $16\,\mathrm{kHz}$.
At each observation step, the policy receives the most recent 16,000 samples, corresponding to a 1.0-s audio window, as
$\mathbf{A}_{\mathrm{raw}}\in\mathbb{R}^{32\times16000}$.
The acoustic observation is updated at 10 Hz.

In simulation, acoustic propagation is recomputed independently at each acoustic update using the current source position.
To construct a temporally continuous waveform from these independently generated segments, only the newly generated portion is appended to a continuous audio stream, and adjacent segments are connected using a 128-sample cross-fade.
The raw-audio observation is extracted from the most recent 1.0 s of this stream, with missing samples at the beginning of an episode zero-padded.

Before being fed to the policy, the waveform is normalized using a channel-wise mean and a standard deviation shared across all 32 channels.
Using a shared standard deviation avoids independently rescaling the microphone channels and therefore preserves relative level differences across microphones that may provide cues for source localization.

\paragraph{Raw-audio encoder}
The waveform is encoded by a four-layer strided 1D convolutional network.
Each block consists of a bias-free 1D convolution, Group Normalization with at most 16 groups, and a GELU activation.
No convolutional padding is used.
The architecture is summarized in Table~\ref{tab:raw_audio_encoder}.
For an input of shape $(B,32,16000)$, the encoder produces a feature map of shape $(B,512,123)$ and contains approximately 1.90 million parameters.
All encoder weights are trained from scratch.

\begin{table}[t]
\centering
\scriptsize
\caption{Architecture of the raw-audio encoder.}
\label{tab:raw_audio_encoder}
\begin{tabular}{ccccc}
\hline
Layer & Output channels & Kernel size & Stride & Output length \\
\hline
1 & 128 & 16 & 8 & 1999 \\
2 & 256 & 8 & 4 & 498 \\
3 & 512 & 4 & 2 & 248 \\
4 & 512 & 4 & 2 & 123 \\
\hline
\end{tabular}
\end{table}

\paragraph{Integration with ACT}
For ACT, the encoder output is adaptively average-pooled along the temporal dimension from 123 to 98 frames and projected to the 512-dimensional transformer hidden space using a $1\times1$ convolution.
The resulting 98 vectors are treated as acoustic tokens, augmented with learnable positional embeddings, and appended to the visual tokens before being passed to the transformer encoder.
The number of acoustic tokens is set to 98 to match S2A2, in which the acoustic spatial-map and spectrogram encoders each produce $7\times7=49$ tokens.
As in the standard ACT configuration, the raw-audio features are not provided to the VAE encoder, which operates only on the robot state and action sequence.

\paragraph{Integration with Diffusion Policy}
For Diffusion Policy, the raw-audio encoder is applied independently to each observation in the observation history.
Temporal mean pooling and maximum pooling are applied to the resulting feature map and concatenated into a 1024-dimensional vector.
This vector is projected to 128 dimensions by a linear layer followed by a ReLU activation and is concatenated with the other observation features used as the global conditioning input of the diffusion model.
The 128-dimensional audio representation matches the combined dimensionality of the acoustic spatial-map and spectrogram features in S2A2, which contribute 64 dimensions each.
With the standard observation history of two steps, raw-audio features from both observations are included in the global conditioning vector.

\paragraph{Training and evaluation}

Except for the acoustic representation and encoder described above, the training and evaluation settings are kept identical to the corresponding S2A2 experiments.
ACT uses a batch size of 8, and Diffusion Policy uses a batch size of 32.
The localization and identification tasks are trained for 100,000 steps using 100 demonstration episodes, whereas the L\&I and exploratory tasks are trained for 200,000 steps using 200 demonstration episodes.
Each policy--task combination is trained with three random seeds and evaluated over 100 episodes per seed.

By matching the number of acoustic tokens in ACT and the acoustic feature dimensionality in Diffusion Policy to those of S2A2, this baseline is designed to isolate the effect of the acoustic representation itself rather than differences in the multimodal fusion interface.

\subsubsection{Hyperparameters in Policy Network Training}
Table~\ref{tab:hyperparams} lists the main training hyperparameters for each policy network.
The number of training steps is generally 100,000 steps; for the exploratory task and the L\&I task experiments, it is set to 200,000 steps to accommodate these more complex tasks.

\begin{table*}[t]
\centering
\scriptsize
\setlength{\tabcolsep}{4pt}
\caption{Training hyperparameters for each policy network. `---' indicates not applicable.}
\label{tab:hyperparams}
\begin{tabular}{l c c c c}
\hline
Item & ACT & Diffusion Policy & VQ-BeT & $\pi_0$ \\
\hline
Optimizer & AdamW & Adam & Adam & AdamW \\
Learning rate & $1.0\times10^{-5}$ & $1.0\times10^{-4}$ & $1.0\times10^{-4}$ & $2.5\times10^{-5}$ \\
Weight decay & $1.0\times10^{-4}$ & $1.0\times10^{-6}$ & $1.0\times10^{-6}$ & $1.0\times10^{-2}$ \\
LR scheduler & None & cosine & warmup only & cosine decay \\
Warmup steps & --- & 500 & 500 & 1,000 \\
Decay steps & --- & --- & --- & 30,000 \\
Final LR & --- & --- & --- & $2.5\times10^{-6}$ \\
Batch size & 8 & 32 & 32 & 4 \\
Observation history & 1 & 2 & 5 & 1 \\
Action horizon & 100 & 16 & 15 & 50 \\
Execution steps per inference & 1 & 8 & 5 & 50 \\
Image normalization & mean/std & mean/std & identity & identity \\
State normalization & mean/std & min/max & min/max & mean/std \\
Action normalization & mean/std & min/max & min/max & mean/std \\
Precision & FP32 & FP32 & FP32 & BF16 mixed \\
\hline
\end{tabular}
\end{table*}

\subsection{Hyperparameters for Audio Processing}
\label{sec:appendix_audio_params}
Table~\ref{tab:audio_params} lists the audio processing parameters used in simulation.
Under the default condition, four microphone arrays are used, and the simulation runs at 30 FPS while the acoustic representation is updated every three frames, yielding an update rate of 10 Hz.
Both the acoustic spatial map and the spectrogram are resized to $224\times224$ pixels before being fed to the policy network.
The NMF in Spotforming is iterated for 40 steps only at the first time step; at subsequent time steps it is terminated after 15 steps using a warm start.
The audio analysis window is 1.0 s (16,000 samples) and is updated every 0.1 s, so adjacent windows overlap by 0.9 s.
The STFT uses a Hann window with an FFT length of 512 and a hop size of 256, which yields 257 frequency bins and 64 time frames for a 16,000-sample window.
In simulation, the signal history is initialized to zero at the beginning of each episode, and the previous acoustic representation is reused at observation frames in which the representation is not updated.
On the real robot, the most recent segment of a ring buffer of at most 1 s is analyzed, and a segment shorter than 1 s is used immediately after the start.
\begin{table}[t]
\centering
\scriptsize
\caption{Audio processing parameters used in simulation.}
\label{tab:audio_params}
\begin{tabular}{ll}
\hline
Item & Value \\
\hline
Sampling rate & 16 kHz \\
STFT FFT length & 512 \\
STFT overlap & 50\% \\
Audio window & 1.0 s \\
Audio update rate & 10 Hz \\
Number of arrays $n$ & 4 \\
Microphones per array & 8 \\
Microphone array radius & 3.5 cm \\
Array placement radius & 30 cm \\
Room size & $10\,\mathrm{m} \times 10\,\mathrm{m} \times 3\,\mathrm{m}$ \\
Reflection order & 3 \\
MUSIC source number & 3 \\
NMF components & 50 \\
NMF threshold & $1.6\times10^{-3}$ \\
NMF iterations (initial) & 40 \\
NMF iterations (warm start) & 15 \\
NMF tolerance & $2.0\times10^{-2}$ \\
NMF beta divergence & 2 \\
NMF max mask ratio & 0.2 \\
Peak selection & local maximum \\
Number of peaks & 1 \\
DAS beamforming normalization & by mic count \\
Spectrogram frequency range & 0--8000 Hz \\
Spectrogram normalization & min--max \\
Distance attenuation $d_{\mathrm{floor}}$ & $10.0\,\mathrm{m}$ \\
Distance attenuation $\gamma$ & $1.0$ \\
\hline
\end{tabular}
\end{table}

\subsection{Differences between Simulation and Real-World Experiments}
\label{sec:appendix_sim_real}
The task definitions and the structure of the acoustic processing are common to simulation and the real robot; only the hardware-dependent settings, which depend on the equipment, placement, and recording characteristics, are changed as described below and summarized in Table~\ref{tab:sim_real}.
\begin{table}[t]
\centering
\scriptsize
\setlength{\tabcolsep}{4pt}
\caption{Hardware-dependent configuration differences between simulation and the real robot.}
\label{tab:sim_real}
\begin{tabular}{p{0.36\textwidth} p{0.27\textwidth} p{0.31\textwidth}}
\hline
Item & Simulation & Real robot \\
\hline
Sampling rate / audio window / update rate & 16 kHz / 1.0 s / 10 Hz & same \\
Array placement & 4 arrays on a circle of radius 0.3 m & 4 arrays at the corners of a 1.2 m $\times$ 0.6 m rectangle \\
Array-center height above working plane & $0.10$ m & $0.14$ m \\
MUSIC source number & 3 & 1 \\
DOA normalization & log-transformed and normalized over azimuth & relative dB (60th-percentile noise floor, 15 dB dynamic range), followed by normalization over frequency bins \\
Map extent & 0.6 m $\times$ 0.6 m & 1.4 m $\times$ 1.4 m \\
Distance attenuation $d_{\mathrm{floor}}$ / $\gamma$ & 10.0 m / 1.0 & 0.0 m / 0.0 (disabled) \\
Exponent applied to each channel before summation & 1 & 4 \\
Gaussian filter before peak extraction & $\sigma=1$ pixel & $\sigma=1$ pixel \\
Peak selection & local maximum & global maximum \\
Spectrogram band / normalization & 0--8 kHz / min--max & 0--4 kHz / 2nd--98th percentile \\
ACT temporal ensemble / executed steps per inference & yes ($0.01$) / 1 & no / 30 \\
\hline
\end{tabular}
\end{table}
In the real robot experiments, the same pipelines as in simulation are used for the acoustic spatial map and the spectrogram.
However, to account for the microphone placement, the robot configuration, and the noise characteristics present in real recordings, we introduced the following differences.

\paragraph{Robot and camera inputs.}
In simulation, we target the Franka Panda and use a 9-dimensional state and a 9-dimensional position control command. The camera inputs are the front and side RGB images rendered in Genesis and saved at $224\times224$ pixels.
In contrast, the real robot experiments use a 7-dimensional state and action comprising the six joints of the right arm of ILOHA and the gripper opening, and feed RGB images obtained from a top-down view and a RealSense camera mounted on the end-effector to the policy network.
Specifically, a $1280\times720$ pixel image is acquired from the top-down RealSense, and a $640\times480$ pixel region at the lower-center of that image is cropped. This processing enlarges the workspace region of the RealSense image for use as input.
A $640\times480$ pixel image is acquired from the end-effector RealSense.
The two $640\times480$ pixel images obtained from the top-down and end-effector cameras are each resized to $224\times224$ pixels and used as inputs to the vision pipeline.

\paragraph{Acoustic input and microphone placement.}
In simulation, Pyroomacoustics is used to model the sound sources and room reflections and to generate synthetic sound. We use four circular 8-channel arrays of radius $0.035\,\mathrm{m}$, with the center of each array placed on a circle of radius $0.3\,\mathrm{m}$.
In the real robot experiments, real recordings are made at 16 kHz from four circular 8-channel arrays of the same specification, with the center of each array placed at the four corners of a $1.2\,\mathrm{m}\times0.6\,\mathrm{m}$ rectangle.

\paragraph{MUSIC and DOA normalization.}
The number of sound sources assumed by MUSIC is set to 3 in simulation and 1 on the real robot.
For the normalization of the DOA spectrum, in simulation the MUSIC output is log-transformed and then normalized by the sum over azimuth.
On the real robot, the MUSIC output is converted to relative dB values using the $60$th percentile as the noise floor and a $15\,\mathrm{dB}$ dynamic range, followed by normalization over the frequency bins.

\paragraph{Acoustic spatial map and Spotforming.}
For the coordinate system of the acoustic spatial map, simulation uses a local grid corresponding to the workspace. The real robot uses map coordinates of $1.4\,\mathrm{m}$ square, matched to the rectangular microphone placement.
The distance attenuation parameters are $d_{\mathrm{floor}}=10.0\,\mathrm{m}$ and $\gamma=1.0$ in simulation, whereas on the real robot they are set to $d_{\mathrm{floor}}=0.0\,\mathrm{m}$ and $\gamma=0.0$, effectively disabling the attenuation.
For peak selection in Spotforming, simulation applies an exponent of $1.0$ to the channel-summed acoustic spatial map and adopts local maxima, whereas the real robot uses an exponent of $4.0$ and adopts the point of maximum value.
For the spectrogram, simulation normalizes the $0$--$8000\,\mathrm{Hz}$ range with min--max normalization, whereas the real robot band-limits to $0$--$4000\,\mathrm{Hz}$ and normalizes using the 2nd--98th percentiles.
\paragraph{Array geometry and calibration.}
On the real robot, the centers of the four arrays are placed at the corners of a $1.2\,\mathrm{m}\times0.6\,\mathrm{m}$ rectangle.
This nominal array placement and the microphone layout within each array are used as the geometric model for constructing the acoustic spatial map, with the azimuth offset of each array set to 0.
We do not perform additional extrinsic calibration between the cameras, the robot, and the microphone arrays, nor do we perform phase, gain, or timing calibration between microphones.
Over the distances considered in our manipulation area, an angular localization error of approximately 5 degrees corresponds to a positional displacement of at most about 5 cm.
The policy therefore operated using the nominal array geometry without additional calibration.

\paragraph{ACT inference settings.}
For ACT inference, simulation uses a temporal ensemble with a weighting coefficient of $0.01$ and re-runs inference after every single executed step.
The real robot experiments do not use a temporal ensemble and continuously execute 30 steps from the action chunk obtained in a single inference.

\section{Additional Latent Space Visualizations} 
\label{sec:appendixB}
	This appendix presents additional t-SNE plots for the policy--task combinations that could not be included in Fig.~\ref{fig:fig4} of the main text.
\subsection{Extraction of Latent States}
\label{sec:appendixB1}
    Because each policy has a different structure, the features used as latent states also differ. For each policy, we extracted both a final-layer representation close to the action output and an intermediate-layer representation from within the policy, and visualized each with t-SNE.
    As the final-layer representation, for ACT we used the features immediately before the representation produced by the transformer decoder is projected onto the action sequence.
    For Diffusion Policy, we used the features immediately before the U-Net output is converted into the noise prediction.
    For VQ-BeT, we used the features immediately before the contextual representation produced by the GPT-style policy network is converted into action tokens or action values.
    For $\pi_0$, we used the features immediately before the representation integrating visual, language, and state information is projected onto continuous actions.

    As the intermediate-layer representation, for ACT we used the output of the transformer encoder that encodes the observation sequence.
    For Diffusion Policy, we used the features near the bottleneck connecting the U-Net encoder and decoder.
    For VQ-BeT, we used the output of the middle layer of the transformer blocks contained in the GPT-style policy network.
    For $\pi_0$, we used the final hidden state on the foundation-model side that integrates visual and language features.
    Because the same U-Net is applied multiple times during the denoising process in Diffusion Policy, we used the features obtained at the midpoint of this process as the intermediate-layer representation.

    The features were recorded at each policy inference. Both the final-layer and intermediate-layer representations were averaged along the sequence axis and treated as a single latent vector per inference step.
    Thus, each point in the t-SNE plot corresponds to the features of an entire action sequence generated by one inference.
    For t-SNE, we set the output dimensionality to 2, used PCA initialization, set the learning rate automatically, and fixed all random seeds.
\subsection{Additional Visualization Results by Policy}
\label{sec:appendixB2}
\begin{figure}[t]
		\centering
		\includegraphics[keepaspectratio, width=14cm]{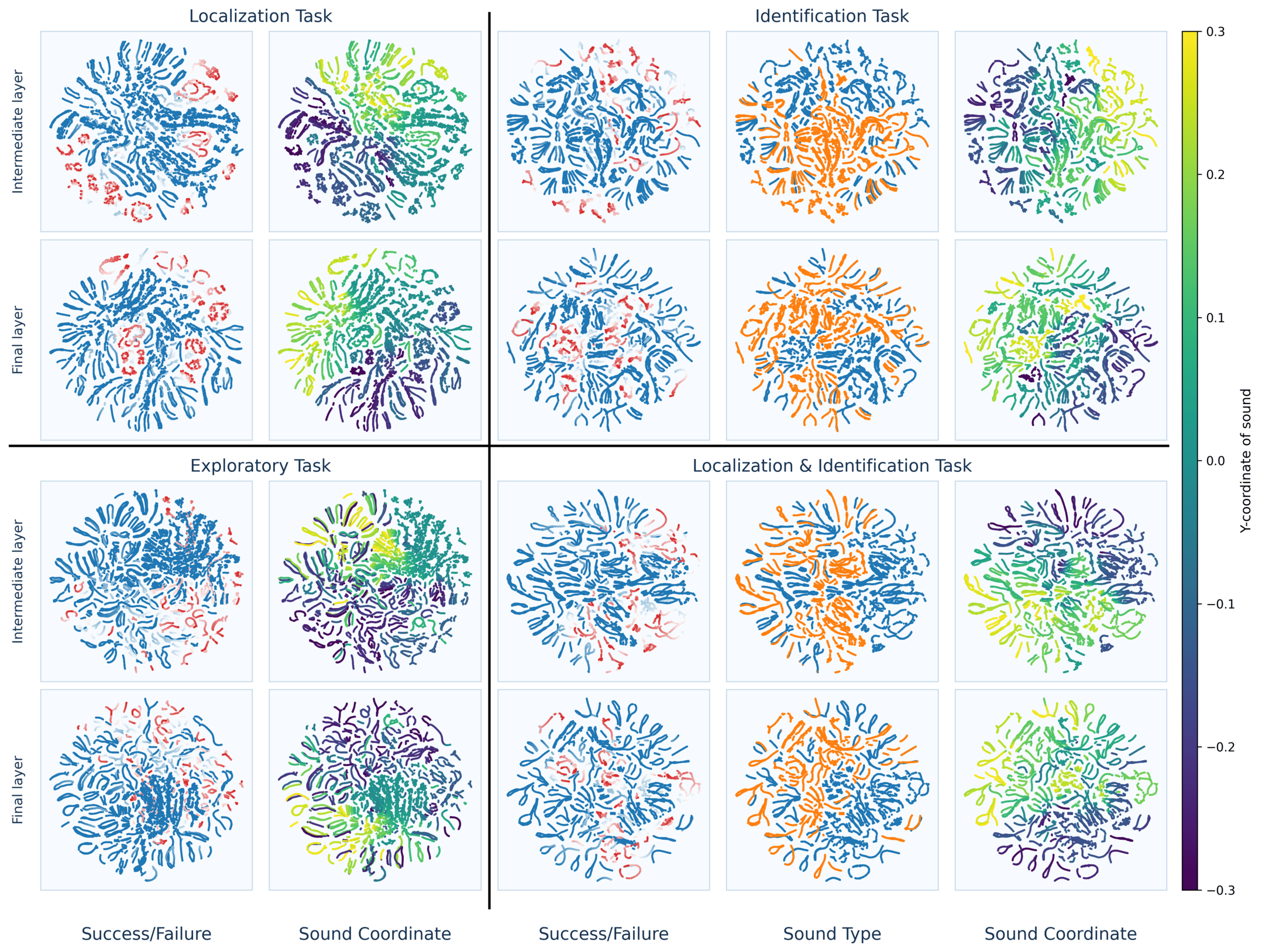}
		\caption{t-SNE visualization of ACT latent representations for the four acoustic-aware manipulation tasks. The four quadrants correspond to Localization, Identification, Exploratory, and L\&I task. In each quadrant, the rows show the same embedded samples colored by different task-related variables: success/failure, sound coordinate, or sound type. 
The columns compare representations from the intermediate layer and the final layer. Each point denotes an observation sample from evaluation trajectories.}
		\label{fig:appendixB_act}
	\end{figure}

\begin{figure}[t]
		\centering
		\includegraphics[keepaspectratio, width=14cm]{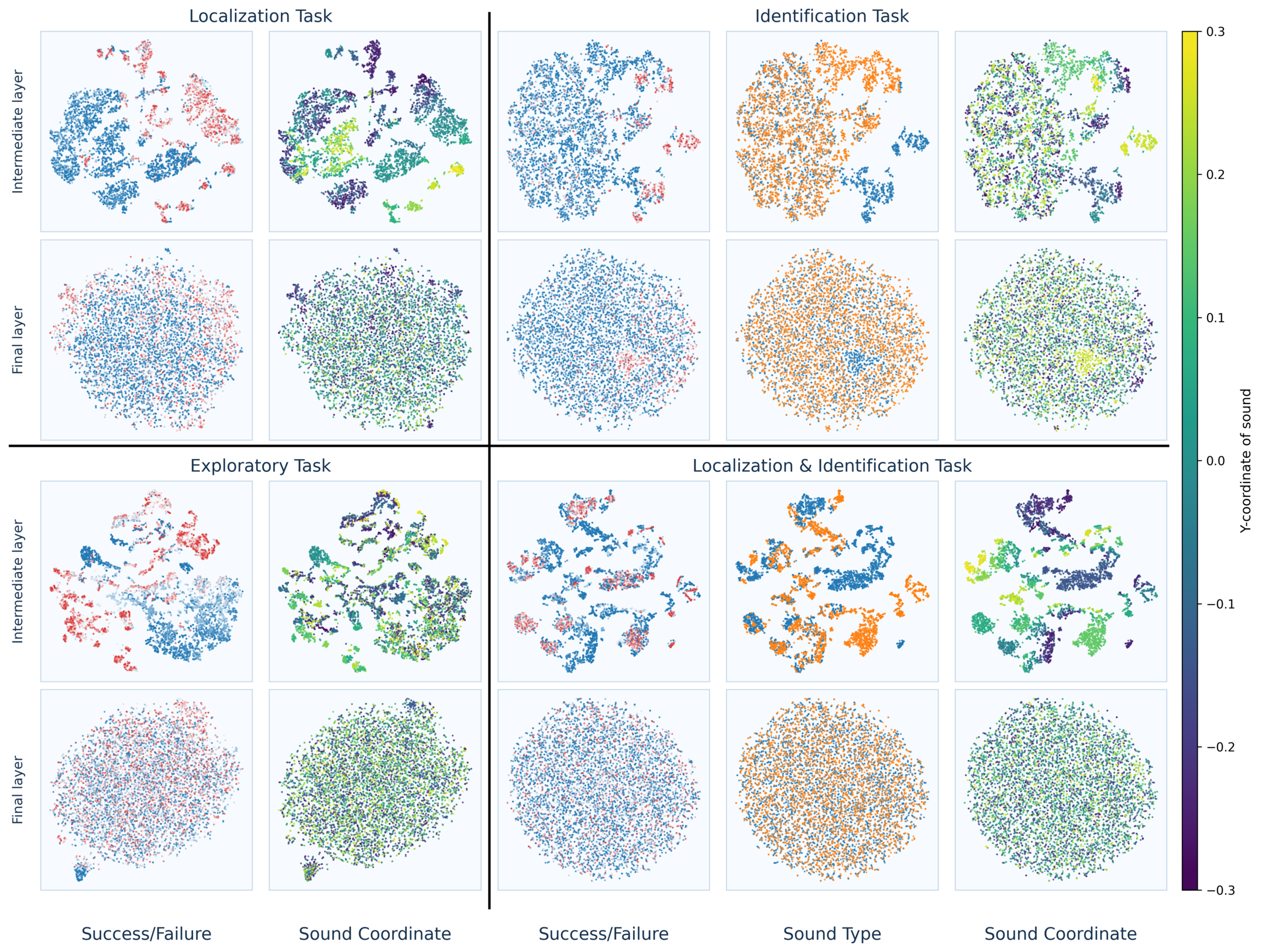}
		\caption{t-SNE visualization of Diffusion Policy latent representations for the four acoustic-aware manipulation tasks. The four quadrants correspond to Localization, Identification, Exploratory, and L\&I task. In each quadrant, the rows show the same embedded samples colored by different task-related variables: success/failure, sound coordinate, or sound type. The columns compare representations from the intermediate layer and the final layer. Each point denotes an observation sample from evaluation trajectories.}
		\label{fig:appendixB_dp}
	\end{figure}

\begin{figure}[t]
		\centering
		\includegraphics[keepaspectratio, width=14cm]{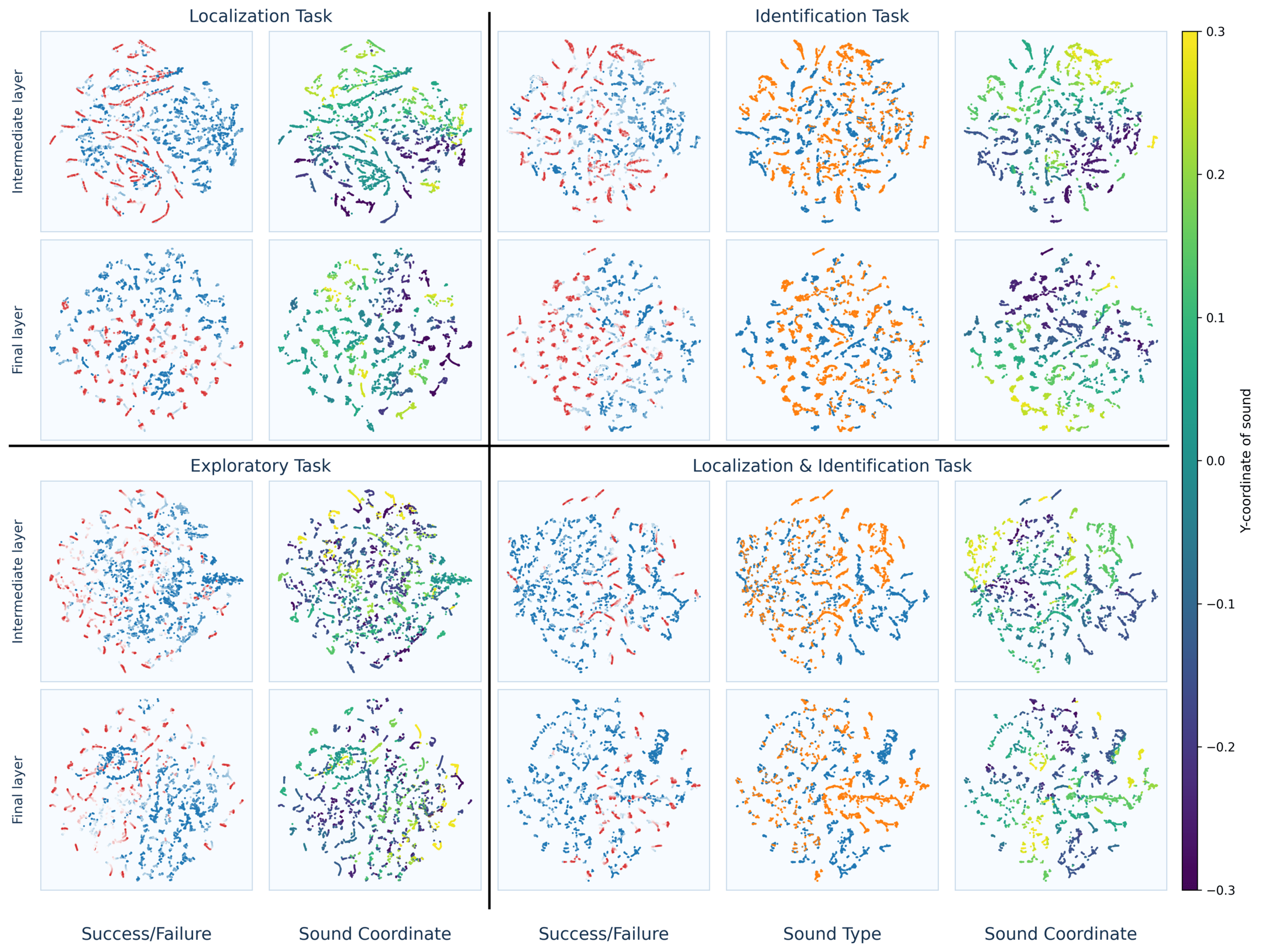}
		\caption{t-SNE visualization of VQ-BeT latent representations for the four acoustic-aware manipulation tasks. The four quadrants correspond to Localization, Identification, Exploratory, and L\&I task. In each quadrant, the rows show the same embedded samples colored by different task-related variables: success/failure, sound coordinate, or sound type. The columns compare representations from the intermediate layer and the final layer. Each point denotes an observation sample from evaluation trajectories.}
		\label{fig:appendixB_vqbet}
	\end{figure}

\begin{figure}[t]
		\centering
		\includegraphics[keepaspectratio, width=14cm]{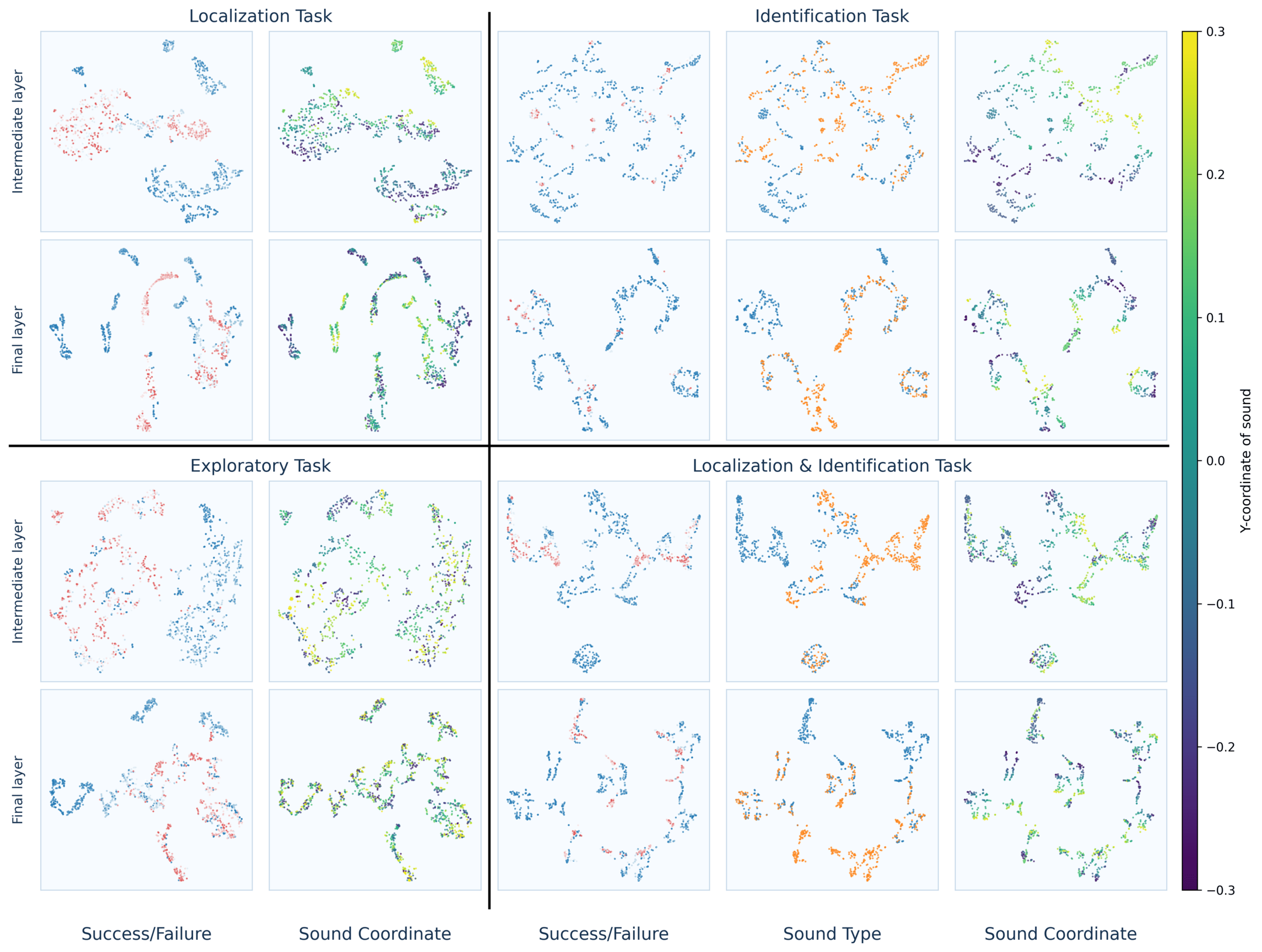}
		\caption{t-SNE visualization of $\pi_0$ latent representations for the four acoustic-aware manipulation tasks. The four quadrants correspond to Localization, Identification, Exploratory, and L\&I task. In each quadrant, the rows show the same embedded samples colored by different task-related variables: success/failure, sound coordinate, or sound type. The columns compare representations from the intermediate layer and the final layer. Each point denotes an observation sample from evaluation trajectories.}
		\label{fig:appendixB_pi0}
	\end{figure}

	Figures~\ref{fig:appendixB_act}--\ref{fig:appendixB_pi0} show the t-SNE plots across all tasks for ACT, Diffusion Policy, VQ-BeT, and $\pi_0$, respectively.
    For each policy network of the S2A2 model, we visualize both the intermediate-layer and final-layer representations on the Localization, Identification, Localization \& Identification (L\&I), and Exploratory tasks.
    Each plot is colored by task success/failure, sound source position, and sound type as labels.
	For task success/failure, data points belonging to successful episodes are shown in blue, while data points belonging to failed episodes are represented with a colormap that transitions from blue at the beginning of the episode, to white in the middle, to red at the end.

\section{Sensitivity Analysis of Audio Representation}
\label{sec:appendix_sensitivity}
\begin{figure}[t]
\centering
\includegraphics[keepaspectratio, width=14cm]{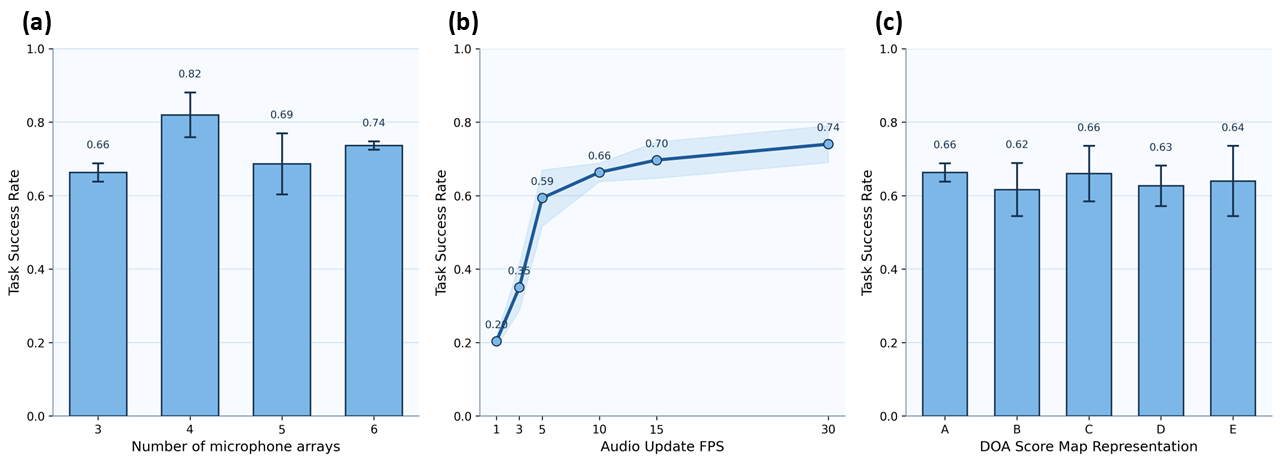}
\caption{Sensitivity analysis of the audio representation. (a) Effect of the number of microphone arrays. (b) Effect of the update FPS of the acoustic information. (c) Comparison of representation methods for the acoustic spatial map.}
\label{fig:appendixA}
\end{figure}

Fig.~\ref{fig:appendixA} shows the sensitivity analysis results for the main hyperparameters of the proposed method.
We used the S2A2 model with ACT for the evaluation.
\paragraph{Effect of the number of microphone arrays.}
Fig.~\ref{fig:appendixA}(a) shows the success rate on the localization task as the number of microphone arrays $n$ is varied. Since the success rate was highest at $n=4$, we adopted this as the standard setting. When the number of arrays is small, the spatial constraints required for DOA estimation are insufficient, whereas when it is too large, the increase in redundant input channels is considered to make policy learning unstable.
\paragraph{Effect of the update FPS of the acoustic information.}
Fig.~\ref{fig:appendixA}(b) shows the results of varying the update frequency of the acoustic spatial map and the spectrogram over the range of 1--30 FPS. Performance improved beyond 5 FPS and roughly saturated from 10 FPS onward. Based on this result, we set the standard update frequency of the acoustic information to 10 FPS.
\paragraph{Comparison of acoustic spatial map representations.}
Fig.~\ref{fig:appendixA}(c) shows the results of varying the representation method of the acoustic spatial map.
A--E in the figure denote differences in the processing applied after projecting the DOA score of each microphone array onto each position in the workspace, as described later in Appendix~\ref{sec:app_doa}.
A is the same as the method described in Section~\ref{sec:doa}, with no additional spatial or temporal processing.
B applies spatial smoothing to the map obtained in A using a Gaussian filter that averages values among neighboring pixels, suppressing local peaks and estimation noise.
C takes a weighted average of the map at each time with the preceding maps, suppressing abrupt changes along the time axis.
D combines the spatial smoothing of B and the temporal smoothing of C, reducing both spatial variation and temporal fluctuation.
E integrates the maps obtained from multiple microphone arrays and converts them into a three-channel feature map consisting of the channel-wise mean of A, a marker indicating the position of maximum response, and a binary mask of the regions with high likelihood.
As a result of the comparison, A and C showed the highest success rates, but C with temporal smoothing did not improve over A without smoothing; we therefore adopted A and used an acoustic spatial map in which the DOA score is directly projected onto the workspace, without additional feature imaging or spatial or temporal smoothing.


In B, a Gaussian filter with $\sigma=1$ pixel (reflect boundary, truncated at $4\sigma$, i.e., an effective $9\times9$ kernel) is applied independently to the 2D map of each array, and each channel is then min--max normalized to 0--255.
In C, an exponential moving average $U_t=(1-\alpha)U_{t-1}+\alpha Q_t$ with $\alpha=0.5$ is applied to the normalized map $Q_t$, and the output is clipped to 0--255.
D applies the spatial smoothing of B followed by the temporal smoothing of C.
E consists of three channels, the mean map over arrays, an $11\times11$-pixel marker centered at the position of maximum response, and a binary mask of the pixels exceeding 0.9 times the maximum of the mean map, each normalized independently to 0--255.


\subsection{Robustness to Acoustic Noise}
\label{sec:appendix_noise}
To evaluate the robustness of the spectrogram obtained by Spotforming to acoustic noise, we compared success rates on the identification task under conditions in which an interfering sound source is added at evaluation time.
The three compared conditions differ only in how the spectrogram is generated; the RGB images, proprioception, and acoustic spatial map are common.
Spot.\ uses the spectrogram obtained by Spotforming, Single uses the waveform of microphone 0 of array 0, and Mean uses the arithmetic mean of the 32 waveforms of the four 8-channel arrays; all conditions apply the same STFT, log-power transform, 0--8 kHz band, min--max normalization, and resizing to $224\times224$ as Spotforming.
For each condition, ACT was trained for 100,000 steps (batch size 8) on 100 noise-free demonstration episodes with three training seeds, and the 100,000-step checkpoint of each seed was evaluated over 100 trials. Noise is added only at evaluation time.
The interfering source is placed at a point on a circle of radius 2 m and height 0.1 m around the workspace center, with its azimuth sampled uniformly and fixed within each episode, and its signal is propagated with the same room model as the target source ($10\times10\times3$ m, reflections up to the third order).
White uses white noise drawn from a standard normal distribution, and Ext.\ loops the other sound of the identification task (sound B when the target emits sound A, and vice versa).
The noise level is specified by the ratio $\eta$ of the RMS of the interfering source waveform to that of the target source waveform before propagation, where $\eta=0$ corresponds to no interfering source.

\begin{table}[t]
\centering
\scriptsize
\setlength{\tabcolsep}{3pt}
\caption{Success rate (\%) on the identification task under acoustic noise (ACT, Mean $\pm$ STD over three seeds, 100 trials each). $\eta$ is the RMS ratio of the interfering source to the target source before propagation. The best value in each column is shown in bold.}
\label{tab:noise}
\begin{tabular}{llccccccc}
\hline
Noise & Spectrogram & $\eta=0$ & 0.25 & 0.5 & 0.75 & 1.0 & 1.25 & 1.5 \\
\hline
White & Spot. & 76.7 $\pm$ 1.2 & \textbf{56.3 $\pm$ 6.4} & \textbf{51.7 $\pm$ 15.2} & \textbf{48.0 $\pm$ 8.9} & \textbf{42.3 $\pm$ 15.6} & \textbf{36.7 $\pm$ 17.2} & 30.3 $\pm$ 19.1 \\
 & Single & 71.0 $\pm$ 5.3 & 43.7 $\pm$ 12.4 & 43.7 $\pm$ 12.7 & 38.0 $\pm$ 13.1 & 28.0 $\pm$ 12.3 & 29.7 $\pm$ 6.7 & 28.3 $\pm$ 6.4 \\
 & Mean & \textbf{81.7 $\pm$ 5.8} & 51.0 $\pm$ 15.6 & 40.0 $\pm$ 8.7 & 32.0 $\pm$ 8.2 & 32.0 $\pm$ 8.5 & 33.0 $\pm$ 2.0 & \textbf{32.7 $\pm$ 9.1} \\
\hline
Ext. & Spot. & 76.7 $\pm$ 1.2 & \textbf{60.0 $\pm$ 4.6} & \textbf{57.7 $\pm$ 14.5} & \textbf{56.7 $\pm$ 7.5} & \textbf{50.3 $\pm$ 7.2} & \textbf{44.3 $\pm$ 10.4} & \textbf{43.7 $\pm$ 14.6} \\
 & Single & 71.0 $\pm$ 5.3 & 53.0 $\pm$ 7.0 & 44.7 $\pm$ 7.4 & 34.0 $\pm$ 4.0 & 33.3 $\pm$ 2.1 & 29.0 $\pm$ 1.0 & 37.3 $\pm$ 3.5 \\
 & Mean & \textbf{81.7 $\pm$ 5.8} & 58.3 $\pm$ 7.0 & 44.0 $\pm$ 4.4 & 38.7 $\pm$ 2.5 & 35.3 $\pm$ 2.3 & 40.0 $\pm$ 3.6 & 35.7 $\pm$ 2.9 \\
\hline
\end{tabular}
\end{table}

Table~\ref{tab:noise} shows the results.
Without noise ($\eta=0$), Mean achieves the highest success rate, whereas with noise, Spotforming achieves the highest success rate for both noise types at all noise levels except $\eta=1.5$ for White.
Spotforming thus trades clean-signal performance for a representation of the acoustic signal that is robust to noise.
Note that this comparison shows robustness to the evaluated noise conditions, and environments with multiple simultaneous task-relevant sources are outside the scope of this work.

\section{Detailed Implementation of the Acoustic Spatial Map Pipeline and the Spectrogram Pipeline}
\label{sec:app_doa} 
This appendix describes the detailed implementation of the acoustic spatial map pipeline and the spectrogram pipeline.
\subsection{Acoustic Spatial Map Pipeline}
    The details of the processing described in Section~\ref{sec:doa} are given below.
	Let the number of arrays be $N$ and the number of microphones per array be $M$.
	Let the STFT of the $M$-channel signal observed at the $c\in\{1,\ldots,N\}$-th array be $\mathbf{x}_{c,f,\tau}\in\mathbb{C}^{M}$. The spatial covariance matrix at frequency bin $f$ is then given by Eq.~(\ref{eq:eq_a}).
	\begin{equation}
	\label{eq:eq_a}
	\mathbf{R}_{c,f}=\frac{1}{T}\sum_{\tau=1}^{T}
	\mathbf{x}_{c,f,\tau}\mathbf{x}_{c,f,\tau}^{\mathrm{H}}
	\end{equation}

	Here, $\mathbb{C}^{M}$ is the $M$-dimensional complex vector space, $\mathbf{R}_{c,f}$ is the spatial covariance matrix at the $f$-th frequency bin of the $c$-th array, $T$ is the number of STFT time frames, and $\mathrm{H}$ denotes the conjugate transpose.
	The MUSIC method obtains the noise subspace $\mathbf{E}_{n,c,f}$ from the eigenvalue decomposition of $\mathbf{R}_{c,f}$ and, based on its orthogonality to the steering vector $\mathbf{a}_{c,f}(\phi)$ for direction $\phi$, computes the spatial spectrum by Eq.~(\ref{eq:eq_b}).
	\begin{equation}
	\label{eq:eq_b}
	P_c(\phi)=\frac{1}{|\mathcal{F}|}\sum_{f\in\mathcal{F}}
	\frac{1}{\mathbf{a}_{c,f}(\phi)^{\mathrm{H}}
	\mathbf{E}_{n,c,f}\mathbf{E}_{n,c,f}^{\mathrm{H}}
	\mathbf{a}_{c,f}(\phi)}
	\end{equation}

	Here, $\mathbf{E}_{n,c,f}$ is the matrix whose columns are the eigenvectors spanning the noise subspace, the subscript $n$ denotes the noise subspace, and $\mathcal{F}$ is the set of frequency bins.

	The log-transformed and normalized spatial spectrum per azimuth is taken as the direction-of-arrival (DOA) score, which is projected onto a $224\times224$ two-dimensional plane corresponding to the plane of the workspace where the microphone arrays lie, generating an $N$-channel acoustic spatial map.
	Let the image row index be $i$ and the column index be $j$, the point in the workspace corresponding to the image coordinate $u=(i,j)$ be $p(u)$, and the array center be $m_c$. The value of each channel is then computed by Eq.~(\ref{eq:eq_c}).
	\begin{equation}
	\label{eq:eq_c}
	S_c(u)=\frac{\bar{P}_c(\psi(p(u),m_c))}
	{\left(\|p(u)-m_c\|_2+d_{\mathrm{floor}}\right)^\gamma}
	\end{equation}

	Here, $\bar{P}_c$ is the normalized DOA score, $\psi(p(u),m_c)$ is the azimuth angle from $m_c$ toward $p(u)$, $\|\cdot\|_2$ is the Euclidean distance, and $d_{\mathrm{floor}}$ and $\gamma$ are parameters that stabilize the distance attenuation.
	Since the DOA score depends only on the azimuth, the same value is assigned to all points along the same azimuth.
	To suppress values far from the workspace, we introduced an attenuation term that depends on the distance from the array center.

\paragraph{Geometry and azimuth grid.}
The speed of sound is set to $343.0\,\mathrm{m/s}$.
Within each array, the relative position of microphone $m\in\{0,\ldots,7\}$ from the array center is $0.035(\cos(2\pi m/8),\sin(2\pi m/8),0)\,\mathrm{m}$, i.e., microphone 0 lies in the $+x$ direction and the index increases counterclockwise when viewed from above.
On the real robot, the azimuth is searched over $[-\pi,\pi)$ at 1-degree intervals (360 directions), and the azimuth of each map pixel is assigned to the nearest search direction; in simulation, the default azimuth grid of Pyroomacoustics~\citep{pyroomacoustics} is used.
No diagonal loading is added to the covariance matrix.

\paragraph{Coordinate correspondence.}
In simulation, the source position $p_{\mathrm{robot}}$ in the robot coordinate frame is mapped to the acoustic room as $p_{\mathrm{room}}=p_{\mathrm{robot}}+(4.5,5.0,0.0)\,\mathrm{m}$, so that the workspace center $(0.5,0,0)$ corresponds to $(5,5,0)$ in the room.
The center of array $c\in\{0,\ldots,N-1\}$ is placed at $(5+0.3\cos\theta_c,\,5+0.3\sin\theta_c,\,0.1)\,\mathrm{m}$ with $\theta_c=\pi(4c-N-3)/(2N)$.
Pixel $(i,j)$ (0-based) of the $224\times224$ map corresponds to $x_{\mathrm{room}}=4.7+0.6\,i/224$ and $y_{\mathrm{room}}=4.7+0.6\,j/224$, i.e., the map covers a $0.6\,\mathrm{m}\times0.6\,\mathrm{m}$ region with rows along $+x$ and columns along $+y$.

On the real robot, the four array centers are represented in the acoustic-map coordinate system as $(5.6,5.3,0.14)$, $(5.6,4.7,0.14)$, $(4.4,4.7,0.14)$, and $(4.4,5.3,0.14)\,\mathrm{m}$ around the center $(5,5)$.
Pixel $(i,j)$ corresponds to $x_{\mathrm{room}}=4.3+1.4\,j/223$ and $y_{\mathrm{room}}=5.7-1.4\,i/223$, covering a $1.4\,\mathrm{m}\times1.4\,\mathrm{m}$ region.
The acoustic spatial map represents positions only in the horizontal plane and does not encode source height.

\paragraph{Normalization of the map.}
Each channel of the projected score $S_c$ is linearly normalized to 0--255 using the minimum and maximum within the current map and stored as an 8-bit image (a constant channel is set to 127.5); this normalization is separate from the normalization of the DOA spectrum over azimuth and from the input normalization of each policy.

\subsection{Spectrogram Pipeline}
\label{sec:app_spec}
	In this work, we generate a spectrogram with suppressed noise and sidelobes based on Spotforming~\citep{spotforming_nmf}.
	First, the integrated map over all arrays is computed as $S_{\mathrm{sum}}(u)=\sum_{c=1}^{N} S_c(u)$. After applying a Gaussian filter to $S_{\mathrm{sum}}$, peak extraction is performed to select source candidates.
	Let the azimuth of the source candidate as seen from each array be $\hat{\theta}_c$. At each array, delay-and-sum beamforming is performed according to $\hat{\theta}_c$ to compensate for the time-of-arrival differences of the sound.

	\begin{equation}
	\label{eq:eq_e}
	y_c(t)=\frac{1}{M}\sum_{m=1}^{M}
	x_{c,m}(t-\tau_{c,m}(\hat{\theta}_c)).
	\end{equation}
	Here, $t$ is time, $x_{c,m}(t)$ is the signal of the $m$-th microphone of the $c$-th array, and $\tau_{c,m}(\hat{\theta}_c)$ is the relative delay for sound arriving from azimuth $\hat{\theta}_c$ to reach each microphone.
	Next, we construct a nonnegative matrix $\mathbf{V}\in\mathbb{R}_{\ge0}^{F_\textrm{b}\times NT_\textrm{b}}$ by concatenating the amplitude spectrograms of the beamforming output $y_c$ of each array along the time axis.
	Here, $F_\textrm{b}$ is the number of frequency bins and $T_\textrm{b}$ is the number of time frames per array. We decompose $\mathbf{V}$ with 50-component NMF as in Eq.~(\ref{eq:eq_f}).

	\begin{equation}
	\label{eq:eq_f}
	\mathbf{V}\approx \mathbf{W}\mathbf{H},\quad
	\mathbf{W}\ge0,\ \mathbf{H}\ge0
	\end{equation}

	Here, $\mathbb{R}_{\ge0}$ is the set of nonnegative reals, $\mathbf{W}\in\mathbb{R}_{\ge0}^{F_{\textrm{b}}\times R}$ is the basis matrix, $\mathbf{H}\in\mathbb{R}_{\ge0}^{R\times N T_{\textrm{b}}}$ is the activation matrix, and $R=50$ is the number of NMF components.
	Let $\mathbf{H}^{(c)}\in\mathbb{R}_{\ge0}^{R\times T_\textrm{b}}$ be the activation matrix split into the time interval of each array. We estimate the components that are commonly strong across all arrays by Eq.~(\ref{eq:eq_g}).
	\begin{equation}
	\label{eq:eq_g}
	\mathbf{Z}=\min_{c}\mathbf{H}^{(c)}
	\end{equation}

	Here, $\min_c$ is the operation that takes the minimum over arrays for each NMF component and each time.
	Let $\tilde{\mathbf{Z}}$ be the matrix obtained by normalizing $\mathbf{Z}$ by its maximum value, and construct a binary mask $\mathbf{B}$ with threshold $\delta=1.6\times10^{-3}$ as in Eq.~(\ref{eq:eq_h}).
	\begin{equation}
	\label{eq:eq_h}
	\mathbf{B}_{r,\tau}=
	\begin{cases}
	1 & \tilde{\mathbf{Z}}_{r,\tau}>\delta,\\
	0 & \mathrm{otherwise}
	\end{cases}
	\end{equation}

	Here, $r$ is the index of the NMF component, $\tau$ is the index of the time frame in the NMF activation matrix, and $\mathbf{B}_{r,\tau}=1$ indicates that the component is retained in the reconstruction.
	To prevent the mask from spreading too widely, when the nonzero ratio of $\mathbf{B}$ exceeds 0.2, we adaptively raise the threshold so that only the top 20\% of activations are retained.
	Finally, the masked amplitude spectrogram corresponding to each array $c$ is reconstructed by $\mathbf{W}(\mathbf{H}^{(c)}\odot\mathbf{B})$, where $\odot$ denotes the element-wise product.
	Combining the reconstructed amplitude spectrogram with the STFT phase of the beamforming output $y_c$ of each array, we obtain a time-domain waveform for each array via the inverse STFT.
	The resulting reconstructed waveforms are averaged across arrays to obtain a single-channel reconstructed waveform $\hat{y}(t)$.
	From the waveform $\hat{y}(t)$, which reconstructs and integrates the common components extracted by NMF, we generate a noise-suppressed single-channel log-power spectrogram.
	Letting the STFT of $\hat{y}(t)$ be $\hat{Y}_{f,\tau}$, the final spectrogram is computed as in Eq.~(\ref{eq:eq_i}).
	\begin{equation}
	\label{eq:eq_i}
	G(f,\tau)=10\log_{10}\left(|\hat{Y}_{f,\tau}|^2+\epsilon\right)
	\end{equation}
	Here, $f$ is the frequency bin, $\tau$ is the time frame, and $\epsilon=10^{-10}$ is a small positive constant that stabilizes the logarithm.
	The upper and lower bounds of the frequency band used differ between the simulation and real robot evaluations; the specific values are given in Appendix~\ref{sec:appendix_sim_real}.

\paragraph{Peak extraction.}
The integrated map used to select the source candidate is $K(u)=\mathcal{G}_\sigma\!\left(\sum_{c=1}^{N}S_c(u)^{q}\right)$, where $q=1$ in simulation and $q=4$ on the real robot, and $\mathcal{G}_\sigma$ is a Gaussian filter with $\sigma=1$ pixel (reflect boundary, truncated at $4\sigma$, i.e., an effective $9\times9$ kernel), corresponding to about 2.7 mm in simulation and 6.3 mm on the real robot.
This Gaussian filter is used only for candidate selection and is independent of the smoothing of the policy input compared in Appendix~\ref{sec:appendix_sensitivity}.
In simulation, pixels whose value is not smaller than any of their existing 8-neighbors are taken as candidates and the one with the largest response is selected; on the real robot, the pixel with the maximum response over the whole map is selected.
No silence detection by an absolute threshold is applied.

\paragraph{Delay-and-sum beamforming.}
Each microphone signal is transformed by an rFFT over the whole window, multiplied by the complex conjugate of the steering vector computed from the speed of sound of $343.0\,\mathrm{m/s}$ and an elevation of 0, summed over microphones, divided by $M=8$, and transformed back by an irFFT.
The delays are thus implemented as frequency-dependent phase rotations without rounding to integer samples.

\paragraph{NMF.}
We use the NMF implementation of torchnmf with $R=50$ components, $\beta=2$, and a tolerance of 0.02.
At the first time step, at most 40 iterations are run from the default initialization of the library; at subsequent time steps, the two factor matrices are carried over as a warm start and at most 15 iterations are run, with re-initialization when the input shape changes.
The minimum activation over arrays $\mathbf{Z}$ is normalized by its maximum and binarized with the initial threshold $\delta=1.6\times10^{-3}$; when the nonzero ratio exceeds 0.2, the threshold is raised to the larger of $\delta$ and the 80th percentile of the activations.
When the maximum activation is positive but the mask is empty, the component with the maximum activation is retained.

\paragraph{Spectrogram image.}
The reconstructed waveforms of the arrays are zero-padded to the same length, averaged, and truncated to the first 16,000 samples before computing the spectrogram.
The band-limited log-power spectrogram is normalized to 0--255 by min--max normalization in simulation and by the 2nd--98th percentiles on the real robot, clipped, converted to 8-bit, and resized to $224\times224$ by bilinear interpolation, with frequency increasing along the rows and time along the columns.
A spectrogram with zero normalization range (e.g., silence) becomes a constant image with the value 127.5.

\section{Real Robot Setup Details}
\label{sec:app_real}
    This appendix describes the experimental setup for the real robot experiments discussed in Section~\ref{sec:real}.
    As the evaluation policy, we adopted ACT, which showed stable performance on both the L\&I task and the exploratory task in simulation, and compared four conditions: the RGB-only baseline, w/o Spec., w/o Spat., and S2A2.
    For each task, we collected 100 episodes of demonstration data by human teleoperation; the same demonstrations were shared across the four conditions, and each model was trained independently for 200,000 steps with a single seed and evaluated over 100 trials.
    The environment setup for each task is described below.

	In the L\&I task, we used small Buffalo speakers (BSSP105UBK) as the sound sources.
    Each speaker has external dimensions of $64\times72\times60$\,mm and a mass of about 210\,g.
	To make the speakers easier for the manipulator to grasp, we wrapped yellow-green protective tape around the side of each speaker to form a handle of approximately 40\,mm radius and 25\,mm width.
	The two destination boxes for the speakers are black, with external dimensions of $60\times90\times90$\,mm, and were each fixed 280\,mm from the two ends of the table.

	In the exploratory task, we prepared two visually identical black empty cans.
	Each can is 135\,mm tall, with a diameter ranging from 40\,mm at the top to 66\,mm at the bottom. Only one of the two cans was filled with three metal nuts so that sound is produced when it is grasped and shaken.
	The gray destination box, measuring $95\times145\times230$\,mm, was placed at the center of the table in front of the robot.

\paragraph{Evaluation.}
The time limit per trial is 15 s for the L\&I task and 20 s for the exploratory task.
A trial is counted as successful if the sounding speaker is placed in the box corresponding to its sound (L\&I task) or the sounding can is placed in the box (exploratory task) within the time limit, as judged by the experimenter at the end of each trial; grasp failures, placing the wrong object or placing it in the wrong box, and timeouts are counted as failures.
The success rates of the four conditions are shown in Table~\ref{tab:real}.
\paragraph{Equipment and data collection.}
The frame that holds the robot and the mounting positions of the cameras follow the design of ALOHA 2~\citep{aloha2}.
Three Intel RealSense D455 cameras are installed, of which the top-down camera and the wrist camera of the right arm are used as the RGB inputs to the policy.
Teleoperation is performed with a Meta Quest 3.
The target frequency of the control loop is 60 Hz; demonstrations and both RGB cameras are recorded at 30 FPS, the acoustic representation is updated at a target rate of 10 Hz, and actions are sent at a target rate of 30 Hz during evaluation.
The top-down RGB image is cropped from $1280\times720$ to the region of columns $[320,960)$ and rows $[240,720)$, the wrist image is $640\times480$, and both are resized to $224\times224$ with area interpolation.
Microphone input is acquired in 20 ms chunks, and the audio processing thread analyzes the most recent segment of at most 1 s.
The real-robot setup uses two tables, each measuring $0.60\,\mathrm{m}\times1.20\,\mathrm{m}$.
The manipulation area is approximately $740\,\mathrm{mm}$ in the front--back direction and $700\,\mathrm{mm}$ in the left--right direction.
Because only the right arm of ILOHA is used in the experiments, the effective manipulation area for the active arm is approximately $740\,\mathrm{mm}\times350\,\mathrm{mm}$.
The centers of the four microphone arrays are located $140\,\mathrm{mm}$ above the working plane.
\paragraph{Sound sources and objects.}
The speakers loop the specified WAV file continuously.
The playback volume was set so that the sound is clearly audible, and the same volume was used for demonstration collection and evaluation.
In the exploratory task, the sounding can contains three M3 nuts.
Object initial positions are randomized within the $740\,\mathrm{mm}\times350\,\mathrm{mm}$ manipulation area used by the right arm.

\section{Additional Real Robot Experiment Sequences}
\label{sec:appendixC}
\begin{figure}[t]
		\centering
		\includegraphics[keepaspectratio, width=14cm]{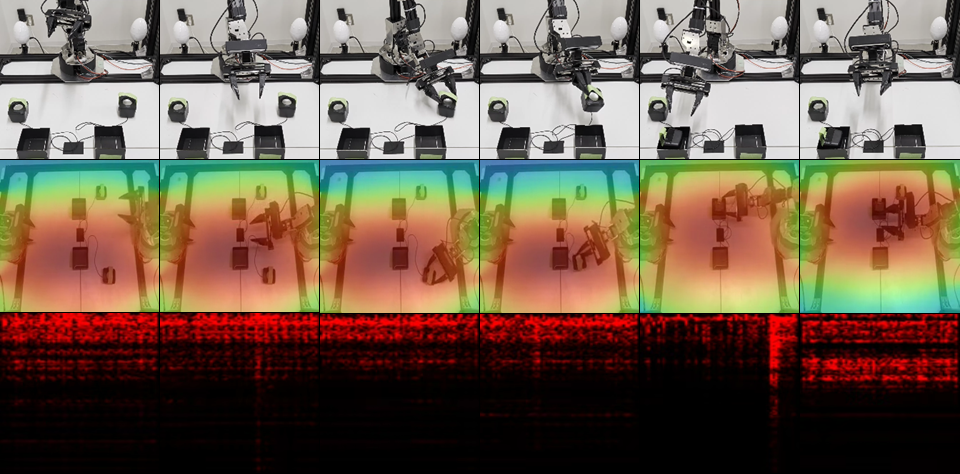}
		\caption{Example real robot sequence of the S2A2 model with ACT performing the L\&I task.}
		\label{fig:real_li_task}
	\end{figure}

\begin{figure}[t]
		\centering
		\includegraphics[keepaspectratio, width=14cm]{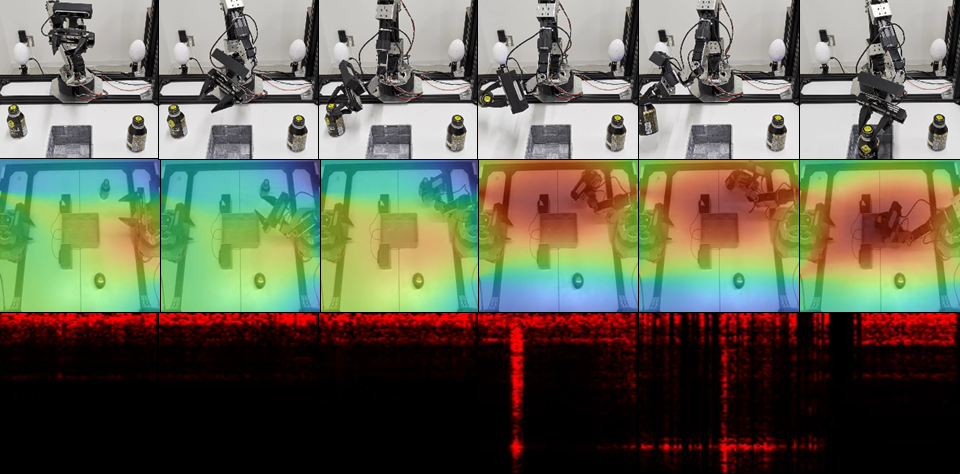}
		\caption{Example real robot sequence of the S2A2 model with ACT performing the exploratory task.}
		\label{fig:real_ex_task}
	\end{figure}

    This appendix presents figures of the real robot experiments that could not be included in the main text, in Figs.~\ref{fig:real_li_task} and~\ref{fig:real_ex_task}.
\section{Detailed results}

\label{sec:results} 
    \paragraph{Performance degradation from mismatched or unnecessary acoustic representations.}
    When an acoustic processing pipeline that did not provide task-relevant information was included, or when a representation required by the task was removed, performance sometimes fell below the vision-only baseline. Examples include the localization task under w/o Spat., where the model receives spectrogram features but lacks the acoustic spatial map needed to identify the sounding object, and Diffusion Policy on the exploratory task under w/o Spat., where using only the spectrogram pipeline degraded performance. These results suggest that, under demonstration data with limited quantity and diversity, policy networks may fail to ignore task-irrelevant acoustic inputs, and such inputs can hinder optimization.


	\textbf{Improvement from leveraging acoustic information.}
	From the results in Table~\ref{tab:results}, the magnitude of improvement from adding acoustic information strongly depends on the policy.
	In particular, ACT and Diffusion Policy showed marked improvements over the baseline by adding acoustic information across all tasks.
	VQ-BeT, on the other hand, showed performance improvements over the baseline across all tasks, but the magnitude of these improvements was consistently smaller than that of ACT and Diffusion Policy.
	A possible cause is that VQ-BeT is designed to quantize actions into discrete codes, which may limit the expressiveness of the actions.
	For the S2A2 model with $\pi_0$, the success rate on the identification task approached that of ACT, which achieved the highest performance, whereas on the exploratory task it was the lowest among all policies.
	Because $\pi_0$ is a large-scale pretrained Vision-Language-Action (VLA) model, an integration of acoustic features that distorts the pretraining distribution may have led to the performance degradation.
	The exploratory task is particularly notable in that the improvement from leveraging acoustic information differs greatly across policies.
	For ACT, conditions that include either acoustic spatial information or acoustic signal information alone substantially exceeded the baseline, and the S2A2 model combining both showed a comparably high success rate.
	Since this task involves selecting the object whose sound was observed after grasping, either acoustic representation---acoustic spatial information or acoustic signal information---can be interpreted as providing the information needed for the conditional branching.
	On the other hand, for Diffusion Policy, VQ-BeT, and $\pi_0$, the performance gains when combined with the S2A2 model were limited.
	In carrying out this relatively long-horizon task involving conditional branching, differences in the action expressiveness of the policies themselves and in the design of how acoustic features are integrated within the multimodal policy are considered to be factors causing the differences across policies.
\section{Simulation Setup Details}
\label{sec:app_simu} 
    We use a Franka Emika Panda robot arm. The proprioception included in the observation is a 9-dimensional vector comprising the end-effector position, the rotation quaternion, and the 2-DoF gripper opening, and the action is a 9-dimensional position control command comprising the 7 joint position targets of the Franka and the 2-DoF gripper position target.
	As the visual information in the observation, we use two $224\times224$ RGB images obtained from the front and side cameras.
	The acoustic simulation uses a $10\,\mathrm{m}\times10\,\mathrm{m}\times3\,\mathrm{m}$ indoor environment and considers reflections up to the third order.
	The microphone arrays are placed at equal intervals on a circle of radius 300\,mm around the workspace center.
	Based on the preliminary experiments shown in Appendix~\ref{sec:appendix_sensitivity}, the number of microphone arrays is set to $N=4$, the visual information is updated at 30 FPS, and the acoustic information at 10 FPS.
	In the exploratory task, one of the two objects is randomly selected per episode, and sound is produced when the movement speed of that object exceeds $0.5\,\mathrm{mm/s}$.
	Two types of sound sources, sound type A and B, are prepared, with A being a bird call and B a buzzer sound. The localization and exploratory tasks use only sound A, while the identification and L\&I tasks use both sounds A and B.
	Demonstrations for imitation learning were collected by generating trajectories with an expert that uses privileged information such as the position of the sound-emitting object and the sound label to specify the target end-effector pose at fixed time intervals, using the damped least-squares inverse kinematics of the Genesis library.
	We collected 100 episodes of demonstrations for the localization and identification tasks, and 200 episodes for the L\&I and exploratory tasks.
	As the training setting, we trained for 100,000 steps on the localization and identification tasks, and for 200,000 steps on the L\&I and exploratory tasks.
	The batch size was 4 for $\pi_0$, 8 for ACT, and 32 for Diffusion Policy and VQ-BeT.

In the Genesis scene, the target objects are cubes with a side of 0.05 m, a density of $50\,\mathrm{kg/m^3}$, and a friction coefficient of 1.5.
Each destination box consists of a $0.1\times0.1\times0.01\,\mathrm{m}$ base plate and four side walls of thickness 0.01 m and height 0.05 m; a single box is placed at $(0.5,0,0)\,\mathrm{m}$ and two boxes at $(0.5,\pm0.15,0)\,\mathrm{m}$.
The front camera is placed at $(2.5,0,1.5)\,\mathrm{m}$ looking at $(0.5,0,0.1)\,\mathrm{m}$ with a field of view of 18 degrees, and the side camera at $(0.5,1.5,1.5)\,\mathrm{m}$ looking at $(0.5,0,0)\,\mathrm{m}$ with a field of view of 20 degrees.

\paragraph{Acoustic simulation.}
The position of the object at each update is used as the source position, and the sound is propagated in the $10\times10\times3\,\mathrm{m}$ room with reflections up to the third order; the directivity of sources and microphones and the wall absorption follow the default settings of Pyroomacoustics.
The source files are 44.1 kHz 16-bit WAV files, which are converted to 16 kHz mono, normalized by the maximum absolute amplitude in the file, and played in a loop.
In the exploratory task, a silent interval of 2 s is placed at the beginning of each episode, and the source signal is passed only while the speed of the selected object is at least $0.5\,\mathrm{mm/s}$; the sound is thus the recorded sound gated by the object motion rather than a physical synthesis of the object vibration.

\end{document}